\documentclass{article} 
\usepackage{iclr2024_conference, times}
\iclrfinalcopy

\usepackage{amsmath,amsfonts,bm}

\def\eqref#1{equation~\ref{#1}}

\def\1{\bm{1}}

\DeclareMathAlphabet{\mathsfit}{\encodingdefault}{\sfdefault}{m}{sl}
\SetMathAlphabet{\mathsfit}{bold}{\encodingdefault}{\sfdefault}{bx}{n}

\usepackage{hyperref}
\usepackage{url}
\usepackage{amsfonts}       %

\usepackage{xcolor}         
\usepackage{stmaryrd}
\usepackage{amsmath}
\usepackage{graphicx}
\usepackage{wrapfig}
\usepackage{caption}
\usepackage{subcaption}
\usepackage{tabularx}
\newtheorem{exmp}{Example}[section]
\usepackage{enumitem}

\title{Improved Variational Bayesian Phylogenetic\\ Inference using Mixtures}

\author{Oskar Kviman\thanks{ Equal contribution.}, Ricky Molén$^*$ \& Jens Lagergren\\
Department of Electrical Engineering and Computer Science\\
KTH Royal Institute of Technology, Science for Life Laboratory\\
Stockholm, Sweden \\
\texttt{\{okviman, rickym, jensl\}@kth.se}
}

\begin{document}

\maketitle

\begin{abstract}
  We present VBPI-Mixtures, an algorithm designed to enhance the accuracy of phylogenetic posterior distributions, particularly for tree-topology and branch-length approximations. Despite the Variational Bayesian Phylogenetic Inference (VBPI), a leading-edge black-box variational inference (BBVI) framework, achieving remarkable approximations of these distributions, the multimodality of the tree-topology posterior presents a formidable challenge to sampling-based learning techniques such as BBVI. Advanced deep learning methodologies such as normalizing flows and graph neural networks have been explored to refine the branch-length posterior approximation, yet efforts to ameliorate the posterior approximation over tree topologies have been lacking. Our novel VBPI-Mixtures algorithm bridges this gap by harnessing the latest breakthroughs in mixture learning within the BBVI domain. As a result, VBPI-Mixtures is capable of capturing distributions over tree-topologies that VBPI fails to model. We deliver state-of-the-art performance on difficult density estimation tasks across numerous real phylogenetic datasets.
\end{abstract}

\section{Introduction}
Phylogenetic inference has a wide range of applications in various fields, such as molecular evolution, epidemiology, ecology, and tumor progression, making it an essential tool for modern evolutionary research. Bayesian phylogenetics allows researchers to reason about uncertainty in their findings about the evolutionary relationship between species.

The posterior distribution over phylogenetic trees given the species data is, however, challenging to infer, since the latent space is a Cartesian product of the discrete tree-topology space and the continuous branch-length space. Furthermore, the cardinality of the tree-topology space grows as a double factorial of the number of species (taxa), making the marginal likelihood computationally intractable in most interesting problem settings.

For over two decades, Markov Chain Monte Carlo (MCMC) approaches have been the go-to approaches for Bayesian phylogenetic analysis, where the MrBayes software \citep{huelsenbeck2001mrbayes} has been particularly popular. However, random walk Metropolis-Hastings MCMC methods \citep{huelsenbeck2001mrbayes, hohna2016revbayes} rely on local operations to explore the tree-topology posterior, a limitation which is known to require long MCMC runs in order to visit  posterior regions which are separated by low-probability tree topologies \citep{whidden2015quantifying}. Sequential Monte Carlo methods for Bayesian phylogeny \citep{bouchard2012phylogenetic, wang2015bayesian, wang2020annealed} have been proposed to avoid these local operations, but the resampling mechanism can filter out important trees in early steps of the algorithm as well as cause degeneracy, necessitating many particles.

More recently, variational inference (VI) has been applied to Bayesian phylogenetics.
In general, VI is often promoted over sampling-based approaches in high-dimensional problems as a variational approximation of the posterior is obtained from optimization, making VI less vulnerable to the curse of dimensionality. However, in practice, it can be challenging to do VI without utilizing sampling. For example, in \cite{koptagelvaiphy} coordinate-ascent update equations are derived in the phylogenetic setting, but these are evaluated using importance sampling. 

\begin{figure}
    \centering
     \begin{subfigure}{0.33\textwidth}\includegraphics[width=\textwidth]{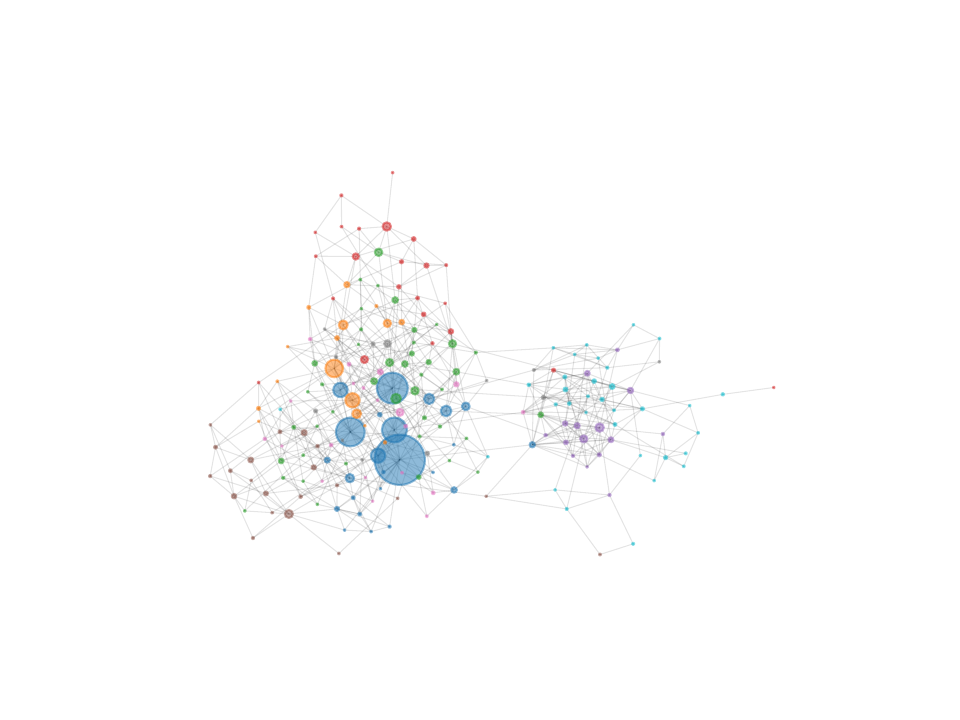}
    \caption{}
    \label{fig:ds4}
    \end{subfigure}~
 \begin{subfigure}{0.33\textwidth}\includegraphics[width=\textwidth]{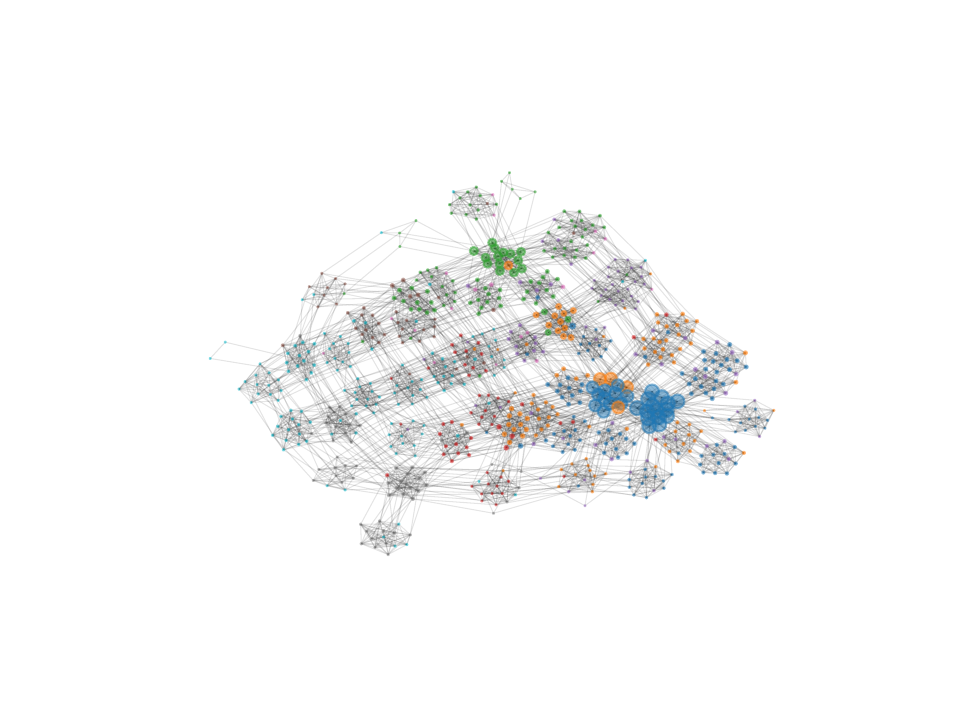}
    \caption{}
    \label{fig:ds7}
    \end{subfigure}~
     \begin{subfigure}{0.33\textwidth}\includegraphics[width=\textwidth]{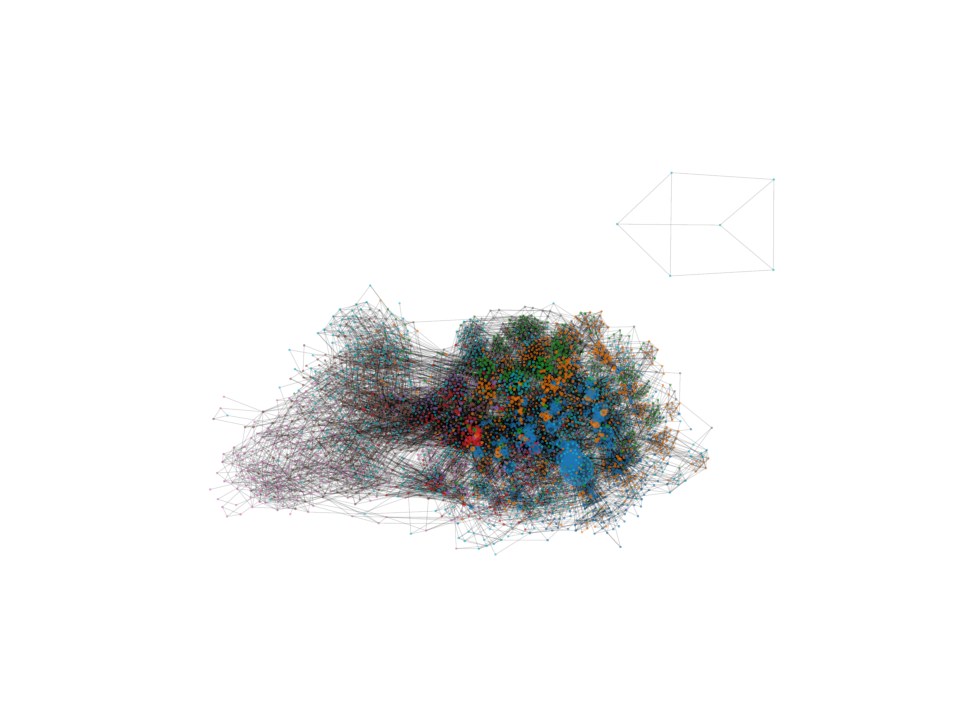}
    \caption{}
    \label{fig:ds8}
    \end{subfigure}
    \caption{Visualization of samples from the tree-topology posterior using a 1,000,000,000 iterations long MCMC run on (a) DS4, (b) DS7 and (c) DS8. Nodes represent unique tree-topologies and are colored based on cluster assignments, illustrating the multimodality of the tree-topology posterior. More details in Sec. \ref{sec:background}.}
    \label{fig:mcmc_topology_space}
\end{figure}

In black-box VI \citep{ranganath2014black}, the gradients are instead taken of Monte-Carlo integrated estimates of the objective, typically the evidence lower bound (ELBO), using samples taken from the variational approximation. This approach has been successfully applied in the Variational Phylogenetic Bayesian Inference (VBPI; \cite{zhang2019variational}) framework, along with its extensions \citep{zhang2020improved, zhang2023learnable}. In these extensions, more complicated branch-length approximations have been proposed using normalizing flows (NFs; \cite{rezende2015variational}) and graph-neural nets (GNNs; \cite{kipf2016semi}). 

BBVI allows the practitioner to learn posterior approximations without deriving update equations or closed-form gradient formulations, the samples are taken from a distribution that is commonly known to concentrate on high-probability regions of the posterior, resulting in a learning procedure that does not sufficiently explore the posterior distribution. This was addressed for continuous distributions in \cite{ruiz2016overdispersed}, where samples were instead drawn from an overdispersed proposal distribution. In discrete hierarchical models, similar to the targets approximated by VBPI, however, insufficient exploration may result in low-level states not being properly modeled. As the tree-topology posterior is known to typically be multimodal with many ``subpeaks'' \citep{whidden2015quantifying} (visualized in Fig. \ref{fig:mcmc_topology_space}), it is thus of significant importance to encourage the tree-topology approximation to explore the posterior. 

We propose VBPI-Mixtures, a novel combination of two recent advances, VBPI and mixture learning in BBVI \citep{kviman2023cooperation}. The mixture components cooperatively explore the tree-topology posterior during learning, and the increased flexibility of the VBPI-Mixtures results in approximations that can model posteriors intractable for the vanilla VBPI (see Sec. \ref{sec:vbpi-mixtures}). 
Using a toy experiment where we design complicated hierarchical categorical target distributions, we first show that the mixture components specialize in different parts of the solution space and achieve smaller Kullback-Leibler divergences to the targets than a single-component approximation which uses more samples. We then apply VBPI-Mixtures on eight popular real datasets, outperforming the state-of-the-art algorithms. The learned tree-topology approximations are then visualized and compared numerically with MCMC ``golden runs'', illustrating the joint exploration of the tree-topology space by the mixture components. Additionally, we derive the VIMCO estimator \citep{Mnih2016-fp} for mixture approximations. Our contributions can be summarized as follows:
\begin{itemize}
    \item We propose VBPI-Mixtures, a novel algorithm for Bayesian phylogenetics.
    \item We show that mixtures of subsplit Bayesian nets (SBNs) can approximate distributions that a single SBN cannot, making a persuasive case for VBPI-Mixtures.
    \item We derive the VIMCO estimator for mixtures.
    \item We visualize a two-component mixture of SBNs on real data, verifying that the components jointly explore the tree-topology space.
    \item Experimentally, we achieve state-of-the-art results on eight popular real phylogenetics datasets, and show that mixtures of SBNs offer more accurate approximations of the tree-topology posterior.
\end{itemize}

\section{Background}
\label{sec:background}
Let $\mathcal{B}=\{b(e):e\in E(\tau)\}$ denote the set of branch lengths for a 
topology, $\tau$, and $X$ is the data. The posterior distribution over leaf-labeled phylogenetic trees,
\begin{equation}
\label{eq:phyloposterior}
    p(\mathcal{B}, \tau|X) = \frac{p(X|\tau, \mathcal{B})p(\mathcal{B}|\tau)p(\tau)}{p(X)},
\end{equation}
is intractable due to its normalizing constant, $p(X)$, i.e., the marginal likelihood. Furthermore, the data, $X=\{X_1,...,X_N\}\in\Omega^{M\times N}$, are observed sequences of length $M$ on the $N$ leaves of the phylogenetic tree. Each entry in $X_{m,n}$ is a character in the alphabet $\Omega$, e.g., $X_{m,n}\in\{A,C,G,T\}$ if DNA sequences are considered.

Although $p(X)$ is intractable, the three terms in the generative model in Eq. \ref{eq:phyloposterior} can be computed efficiently. Typically, $p(\tau)$ is a uniform distribution over the (rooted or unrooted) tree-topology space, and the branch-length prior is an exponential distribution with rate $\lambda$, such that $p(\mathcal{B}|\tau)=\prod_{e\in E(\tau)}\lambda e^{-\lambda b(e)}$. The likelihood, $p(X|\tau, \mathcal{B})$, can be evaluated in linear (in $N$) time using the standard dynamic programming algorithm proposed by \cite{Felsenstein2003-ts}.  

Let a \textit{clade} be a non-empty subset of the set of the $N$ leaf labels, $\mathcal{X}$. A \textit{subsplit} is a partition of this clade into two lexicographically ordered, disjoint clades, while a \textit{split} is simply a root subsplit---a bipartition of $\mathcal{X}$. Furthermore, a primary subsplit pair (PSP) is a subsplit conditioned on a split---a tripartition of $\mathcal{X}$. In Appendix \ref{sec:intro_to_phylo} we additionally give a brief introduction to Bayesian phylogenetic inference for machine learning researchers.

\paragraph{Variational Inference in Bayesian Phylogenetics}
The VI-based approach to Bayesian phylogenetics is to approximate $p(\mathcal{B}, \tau|X)$ using a simpler distribution, $q_{\psi,\phi}(\mathcal{B}, \tau) = q_{\psi}(\mathcal{B}|\tau)q_\phi(\tau)$. Generally, in VI, the approximations are learned by maximizing the evidence lower bound (ELBO),
\begin{equation}
    \mathcal{L}(X) = \mathbb{E}_{q_{\psi,\phi}(\mathcal{B}, \tau)}\left[\log\frac{p(X|\tau, \mathcal{B})p(\mathcal{B}|\tau)p(\tau)}{q_{\psi}(\mathcal{B}|\tau)q_\phi(\tau)}\right],
\end{equation}
implicitly minimizing the Kullback-Leibler (KL) divergence from $p(\mathcal{B}, \tau|X)$ to $q_{\psi,\phi}(\mathcal{B}, \tau)$.

\paragraph{Subsplit Bayesian Networks}
Given a set of tree topologies, $\mathcal{T}$,\footnote{$\mathcal{T}$ is in practice obtained from some efficient tree-topology sampling algorithm, typically UFBoot \citep{Minh2013-mu}.} it is straightforward to form a look-up table of all subsplits in $\mathcal{T}$. The SBN uses the look-up table to define support over possible tree topologies and learns the probabilities of the subsplits in the table via, for example, stochastic optimization. As the look-up table contains probabilities of subsplits, it is referred to as a \textit{conditional probability table} (CPT). When the CPT has been learned, the SBN provides a tractable probability distribution over tree topologies from which it is possible to sample. See \cite{zhang2018generalizing} or \cite{zhang2022variational} for the original, more in-depth accounts of SBNs.

\paragraph{VBPI}
In VBPI \citep{zhang2019variational}, the posterior approximations are learned by maximizing a multi-sample \citep{Burda2015-lr} version of $\mathcal{L}(X)$, 
\begin{equation}
\label{eq:iwelbo}
    \mathcal{L}(X;K) = \mathbb{E}_{\mathcal{B}^k, \tau^k\sim q_{\psi,\phi}(\mathcal{B}, \tau)}\left[\log\frac{1}{K}\sum_{k=1}^K\frac{p(X|\tau^k, \mathcal{B}^k)p(\mathcal{B}^k|\tau^k)p(\tau^k)}{q_{\psi}(\mathcal{B}^k|\tau^k)q_\phi(\tau^k)}\right],
\end{equation}
where $q_\phi(\tau)$ is an SBN with a learnable CPT, $\phi$, and $q_\psi(\mathcal{B}|\tau)$ is a multivariate LogNormal distribution with a diagonal covariance matrix, such that
\begin{equation}
q_\psi(\mathcal{B}|\tau) = \prod_{e\in E(\tau)}q(b(e)|\mu(e, \tau), \sigma^2(e, \tau)).
\end{equation}
Two different parameterizations of $\mu(e, \tau)$ and $\sigma(e, \tau)$ have previously been proposed. The simpler approach is to let $\mu(e, \tau) = \psi^\mu_{e/\tau}$ and $  \sigma(e,\tau)=\psi^\sigma_{e/\tau}$, where $e/\tau$ denotes a split of $\tau$ in edge $e$. The parameters $\psi^\mu_{e/\tau}$ and $\psi^\sigma_{e/\tau}$ are shared among all tree topologies where $e/\tau$ exists, resulting in an amortized mapping from a tree topology to the parameters of $q_\psi(\mathcal{B}|\tau)$. Additional local information about the given $\tau$ can be added into the parameterization of the approximation by using PSPs,
\begin{equation}
    \mu(e, \tau) = \psi^\mu_{e/\tau} + \sum_{i\in e\sslash\tau} \gamma_i^\mu, \quad \sigma(e, \tau) = \psi^\sigma_{e/\tau} + \sum_{i\in e\sslash\tau} \gamma_i^\sigma,
\end{equation}
 where $e\sslash\tau$ denotes the set of PSPs neighboring to the split $e/\tau$, and $\gamma_i^\mu$ is a learnable parameter associated with the $i$-th pair.

\paragraph{Multimodality of the tree-topology posterior}
In \cite{whidden2015quantifying}, modes are referred to as clusters of MCMC samples that are densely grouped in the tree-topology space and have high posterior density compared to their neighbors. They proposed a method for detecting and quantifying peaks by calculating the reversible subtree pruning and regrafting distance between topologies, combined with the MrBayes MCMC posterior probability. By applying this method, it was identified that certain datasets had a high number of modes (e.g., DS1, DS4, DS5, DS6, DS7), which shows the complexity of the tree-topology space. In Fig. \ref{fig:mcmc_topology_space} we visualize the multimodality of the posterior on three datasets.

\paragraph{Mixtures in Black-Box VI} Learning mixtures of approximations in (black-box; \cite{ranganath2014black}) VI \citep{nalisnick2016approximate,morningstar2021automatic,kviman2023cooperation} is a compelling off-the-shelf technique to increase the flexibility of a variational approximation.
Mixtures can be applied to any variational approximation, including an NF-based approximation or one for discrete latent variables, with little overhead. 

The objective function, the ELBO for mixtures, is estimated by sampling from each mixture component in a stratified manner \citep{morningstar2021automatic}, or via multiple importance sampling techniques \citep{kviman2022multiple}, why the objective is often referred to as \textit{MISELBO}. Maximizing MISELBO encourages the mixture components to cooperatively cover the target distribution, which is thought to be the key ingredient for their success in density estimation tasks \citep{kviman2023cooperation}. In the next section, we formulate MISELBO for VBPI and explain why mixtures of SBNs are beneficial for exploring the complex tree-topology space.

\section{Variational Bayesian Phylogenetic Inference using Mixtures}
\label{sec:vbpi-mixtures}
Here we present our proposed method, VBPI-Mixtures. We derive the VIMCO estimator for learning mixtures of SBNs, and show how to combine mixtures of branch length models with an expressive NF model. We start by providing the MISELBO formulation for VBPI with $K$ importance samples,
\begin{equation}
\label{eq:miselbo}
    \mathcal{L}(X;K, S) = \frac{1}{S}\sum_{s=1}^S\mathbb{E}_{\mathcal{B}_s^{1:K}, \tau_s^{1:K}\sim q_{\psi_s,\phi_s}(\mathcal{B}, \tau)}\left[\log\frac{1}{K}\sum_{k=1}^K\frac{p(X|\tau_s^k, \mathcal{B}_s^k)p(\mathcal{B}^k_s|\tau_s^k)p(\tau_s^k)}{\frac{1}{S}\sum_{j=1}^S q_{\psi_j}(\mathcal{B}_s^k|\tau_s^k)q_{\phi_j}(\tau_s^k)}\right].
\end{equation}
To evaluate $\mathcal{L}(X;K, S)$, we approximate the $s$-th expectation by Monte-Carlo integration using simulations from $q_{\psi_s,\phi_s}(\mathcal{B}, \tau)$. 

\begin{figure}
    \centering
    \includegraphics[width=\textwidth]{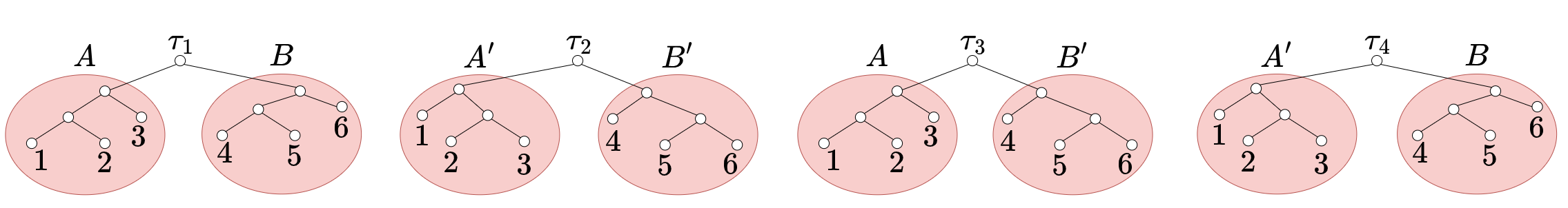}
    \caption{Mixtures of SBNs increase the flexibility of the tree-topology approximation. For instance, they can exactly capture a target distribution that assigns its probability uniformly across $\tau_1$ and $\tau_2$, 
    leaving zero probability to $\tau_3$ and $\tau_4$.
    Meanwhile, this is not possible for a single SBN. See Example \ref{example} for details.}
    \label{fig:four trees}
\end{figure}

\paragraph{Mixtures promote exploration}
Note that minimizing the denominator in Eq. (\ref{eq:miselbo}) corresponds to diversifying the mixture distribution, i.e., promoting the mixture components to jointly explore the latent space. This exploratory behavior is crucial in black-box VI as the ELBO is only evaluated in sampled (visited) latent variables (states). 

More specifically, the samples are proposed from the same distribution that attempts to maximize the ELBO. Consequently, for $S=1$, there is a risk that, for a fixed $K$, less probable regions will not be sufficiently explored during learning and thus will be poorly modeled. In fact, \cite{zhang2019variational} showed that the vanilla VBPI does not benefit from more importance samples during training. Fortunately, the MISELBO objective offers a promising solution, as the mixture components are promoted to spread out and efficiently explore the multimodal phylogenetic posterior.

\paragraph{Mixtures increase the flexibility of the approximations} A mixture of LogNormal pdfs is clearly more flexible than a single LogNormal pdf. For mixtures of SBNs, this is also true but may be less clear. An SBN  constructs a tree by stochastically partitioning the available clades. The partition of a clade is sampled \textit{independently} of the partitions sampled in the other clades. Mixtures of SBNs allow for modeling correlations in the sampling of the partitions, and thus increase the flexibility of the approximation. We explain this feature with a simple example, which trivially generalizes to larger trees and also applies to unrooted trees.

\begin{exmp}
\label{example}
Consider the four rooted tree topologies in Fig. \ref{fig:four trees} where $A,B,A'$ and $B'$ are four different subtrees for the two clades (in red), and a target density, such that $p(\tau_1) = p(\tau_2) = 0.5$. A uniformly weighted mixture of two SBNs can easily approximate $p$ exactly by letting $q_1(\tau_1) = 1$ and $q_2(\tau_2)=1$, resulting in $\frac{1}{2} q_1(\tau_1) + 0 = 0 + \frac{1}{2}q_2(\tau_2) = 0.5$. 
However, as a single SBN, $q$, samples $A$ or $A'$ and $B$ or $B'$ independently, and in order to achieve $q(\tau_1), q(\tau_2)>0$, it will have to assign 
non-zero probability to all four subtrees. 
Specifically, say $q(A)=\alpha$ and, consequently,  $q(A')=1-\alpha$, while $q(B)=\beta$ and, consequently, $q(B')=1-\beta$. 
It follows that $q(\tau_3)= \alpha (1-\beta)$ and $q(\tau_4)= (1-\alpha) \beta$. Finally,
$\alpha \beta = q(\tau_1)=q(\tau_2)=(1-\alpha) (1-\beta )$ implies 
$1=\alpha + \beta$, which, in turn, implies that $q(\tau_3)=\alpha^2$ and $q(\tau_4)=\beta^2$. 
That is, all four trees will either have probability $1/4$ under $q$, or one of $\tau_3$ and $\tau_4$
will have a higher probability than $\tau_1$ and $\tau_2$. So, a single SBN can only yield a distribution that is very different from $p$.
\end{exmp}

The above example exemplifies that there are tree-topology distributions which cannot be modeled using a single SBN, but which can be modeled by a mixture of SBNs. There is no converse example, as a mixture can trivially model a single SBN by letting $q_1(\tau)=q_2(\tau)$ for all $\tau$. In Appendix \ref{sec:DNA-example} we construct an example which shows that these conflicting tree-topology posteriors can indeed occur for real DNA data.

\subsection{VIMCO for Mixtures of Tree-Topology Approximations}
Here we derive the VIMCO estimator of Eq. (\ref{eq:miselbo}). Although the notation in this section is specific for Bayesian phylogenetics, our result is applicable to any mixture approximation. We purposefully follow the derivations in \cite{Mnih2016-fp} closely. 

\subsubsection{Gradient Analysis}
The gradients of Eq. (\ref{eq:miselbo}) are studied first. Let
\begin{equation}
    f(x, \mathcal{B}^k_s, \tau^k_s)=  \frac{p(\mathcal{B}^k_s, \tau^k_s, X)}{\frac{1}{S}\sum_{j=1}^S q_{\psi_j}(\mathcal{B}^k_s| \tau^k_s)q_{\phi_j}(\tau^k_s)},
\end{equation} 
and $\hat{L}^K_s = \log \frac{1}{K}\sum_{k=1}^K f(x, \mathcal{B}^k_s, \tau^k_s)$, where $\mathcal{B}^k_s, \tau^k_s$ are simulated from $ q_{\psi_s, \phi_s}(\mathcal{B}, \tau)$. Note that $f(x, \mathcal{B}^k_s, \tau^k_s)$, and so also $\hat{L}^K_s$, is a function of the SBN parameters for all mixture components, i.e., $\phi_1,...,\phi_S$. However, we omit these as arguments to the function in order to avoid cluttered notation. 

We are interested in the gradient of Eq. (\ref{eq:miselbo}) with respect to the SBN parameters for one of the mixture components, say $i$,
\begin{equation}
\label{eq:grad_i_miselbo}
    \nabla_{\phi_i} \mathcal{L}(X;K, S) = \nabla_{\phi_i} \frac{1}{S}\sum_{s=1}^S\mathbb{E}_{q_{\psi_s,\phi_s}(\mathcal{B}, \tau)}\left[
    \hat{L}^K_s
    \right].
\end{equation}

The full derivations are given in Appendix \ref{app:gradient_derivations} and lead to the following expression
\begin{equation}
\label{eq:self-normalized_gradient}
\begin{split}
     \nabla_{\phi_i} \mathcal{L}(X;K, S) &= \frac{1}{S}\mathbb{E}_{q_{\psi_i,\phi_i}(\mathcal{B}, \tau)}\Big[ \hat{L}^K_i\sum_{k=1}^K\nabla_{\phi_i} \log q_{\phi_i}(\tau_i^{k})\Big] -\\&\quad
    \frac{1}{S}\sum_{s= 1}^S\mathbb{E}_{q_{\psi_s,\phi_s}(\mathcal{B}, \tau)}\Big[\sum_{k=1}^K\tilde{w}^k_s\nabla_{\phi_i}\log \frac{1}{S}\sum_{j=1}^S q_{\psi_j}(\mathcal{B}^k_s| \tau^k_s)q_{\phi_j}(\tau^k_s)\Big].
\end{split}
\end{equation}
We make three important observations, $(i)$ for $S=1$, we retrieve the gradients used to train VBPI, $(ii)$ as the second term is negated, maximizing it corresponds to diversifying the mixture distribution w.r.t. $\phi_i$, and $(iii)$ the first term is merely a scaled (by $1/S$) version of the corresponding term in the $S=1$ case. Connecting to observation $(iii)$, we conclude that extending the VIMCO estimator to $S>1$ cannot be trivially achieved without our derivation provided above. 

Furthermore, the analysis of the gradients of the importance weighted lower bound in \citep{Mnih2016-fp}---using our notation, $\mathcal{L}(X;K)$---applies here, too. That is, the gradients in the second term in Eq. (\ref{eq:self-normalized_gradient}) are multiplied by normalized weights, ensuring that the norm of the weighted sum over all $K$ gradients is not greater than the norm of the largest term in the sum. This means that $\phi_i$ will be updated mainly according to gradients based on simulations scored highly by $f$, while mitigating the impact of gradients from lower-scoring simulations. 

In the first term, on the other hand, all gradients are multiplied by the same  $\hat{L}^K_i$, indicating that the gradients of high-scoring simulations will not receive more weight than low-scoring ones, causing high variance and slow learning.

\subsubsection{The VIMCO Estimator}
\label{sec:vimco}
As concluded above, the second term in Eq. (\ref{eq:self-normalized_gradient}) is well-behaved, and we do not need variance-reduction techniques to use it for learning in practice. The first term, however, requires attention in order to facilitate efficient learning. 

Fortunately, as the first term is merely a scaled version of the corresponding term in the VIMCO estimator when $S=1$, we can directly apply the localized learning signal strategy from \cite{Mnih2016-fp} to obtain the VIMCO estimator for $S\geq 1$,
\begin{equation}
\label{eq:vimco_estimator}
    \begin{split}
     \nabla_{\phi_i} \mathcal{L}(X;K, S) &\simeq \frac{1}{S} \sum_{k=1}^K\hat{L}^K_{i(k|-k)}\nabla_{\phi_i} \log q_{\phi_i}(\tau_i^{k}) -\\&\quad
    \frac{1}{S}\sum_{s=1}^S\sum_{k=1}^K\tilde{w}^k_s\nabla_{\phi_i}\log \frac{1}{S}\sum_{j=1}^S q_{\psi_j}(\mathcal{B}^k_s| \tau^k_s)q_{\phi_j}(\tau^k_s),
\end{split}
\end{equation}
where $\tau^{k}_i, \mathcal{B}_i^{k}\sim q_{\phi_i, \psi_i}(\tau, \mathcal{B})$ and $\tau^{k}_s, \mathcal{B}_s^{k}\sim q_{\phi_s, \psi_s}(\tau, \mathcal{B})$. Here, $\hat{L}^K_{i(k|-k)}$ is the local learning signal for sample $k$, defined as
\begin{equation}
    \hat{L}^K_{i(k|-k)} = \hat{L}^K_{i} - \log\frac{1}{K}\left( 
    \sum_{k'\neq k }f\left(x, \tau^{k'}_i, \mathcal{B}_i^{k'}\right) + \hat{f}\left(x, \tau^{-k}_i, \mathcal{B}_i^{-k}\right) 
    \right),
\end{equation}
with $\hat{f}\left(x, \tau^{-k}_i, \mathcal{B}_i^{-k}\right)$ being an estimator of $f\left(x, \tau^{k}_i, \mathcal{B}_i^{k}\right)$, typically the geometric mean \cite{Mnih2016-fp, zhang2019variational, zhang2020improved}, $    \hat{f}\left(x, \tau^{-k}_i, \mathcal{B}_i^{-k}\right) = \exp\left(\frac{1}{K-1}\sum_{k'\neq k}\log f\left(x, \tau^{k'}_i, \mathcal{B}_i^{k'}\right)\right).$

\section{Experiments}
In Sec. \ref{sec:vbpi-mixtures}, we argued that a single-component approximation will struggle to properly model all parts of the target distribution when learned with black-box VI. Below, in Sec. \ref{sec:exploration}, we experimentally verify this claim, and, furthermore, confirm that mixture components collaborate in order to jointly cover the target density, resulting in more accurate approximations and efficient exploration.

We then, in Sec. \ref{sec:real data}, demonstrate that the increased model flexibility and promotion of exploration translates into better marginal log-likelihood estimates and more accurate tree-topology posterior approximations. We also visualise representations of VBPI-Mixtures on real data. Code for all experiments is provided at \url{https://github.com/Lagergren-Lab/VBPI-Mixtures}.

\subsection{Exploring a Discrete Two-Level Hierarchical Model using Black-Box VI}
\label{sec:exploration}
SBNs are hierarchical models with categorical distributions at each level. Here, we examine how mixtures explore discrete hierarchical target distribution, when learned via black-box VI. 

 \begin{figure}[!t]
    \centering
\begin{subfigure}{0.33\textwidth}
    \includegraphics[width=\textwidth]{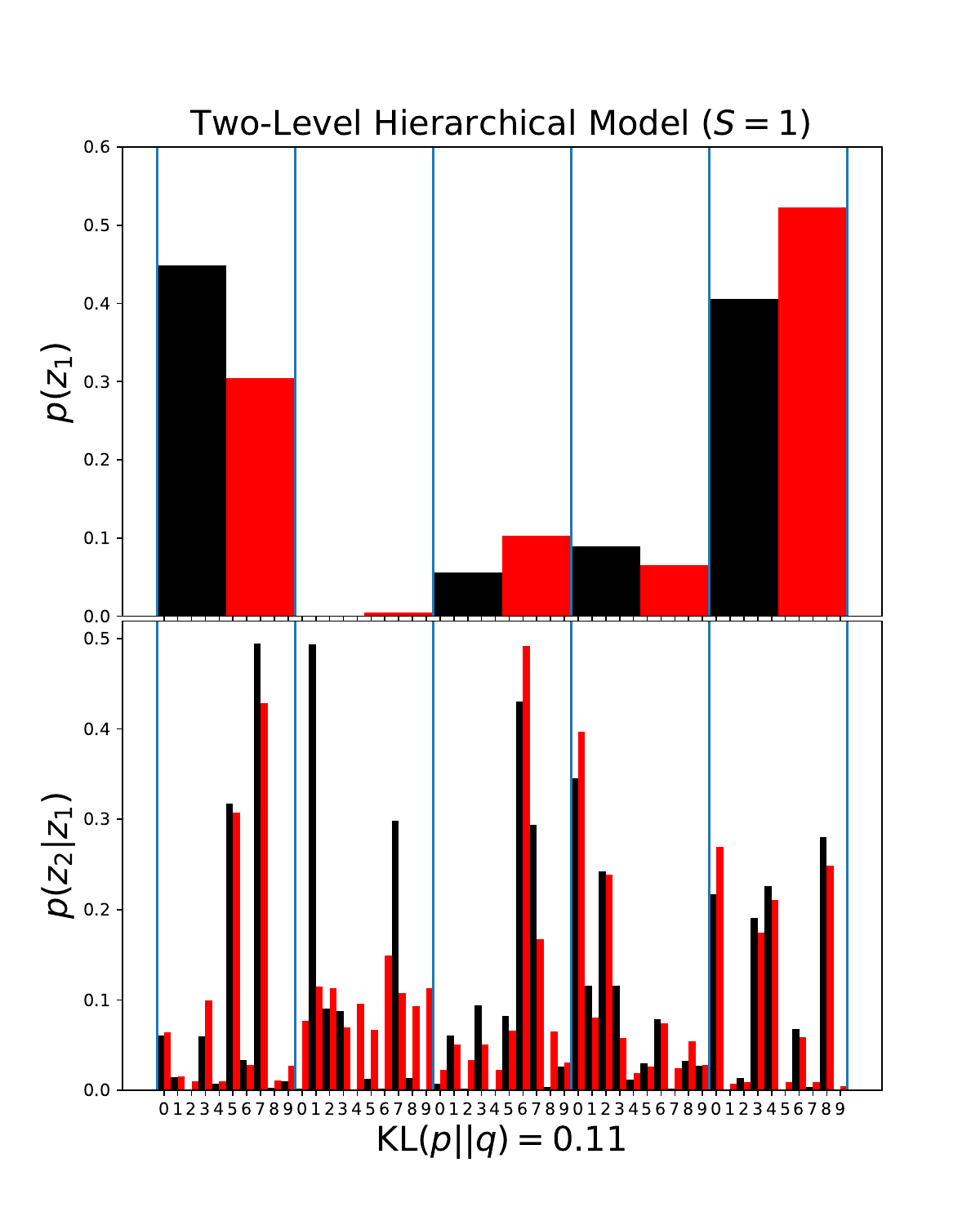}
    \caption{}
    \label{fig:my_label_2}
\end{subfigure}~
\begin{subfigure}{0.33\textwidth}
    \includegraphics[width=\textwidth]{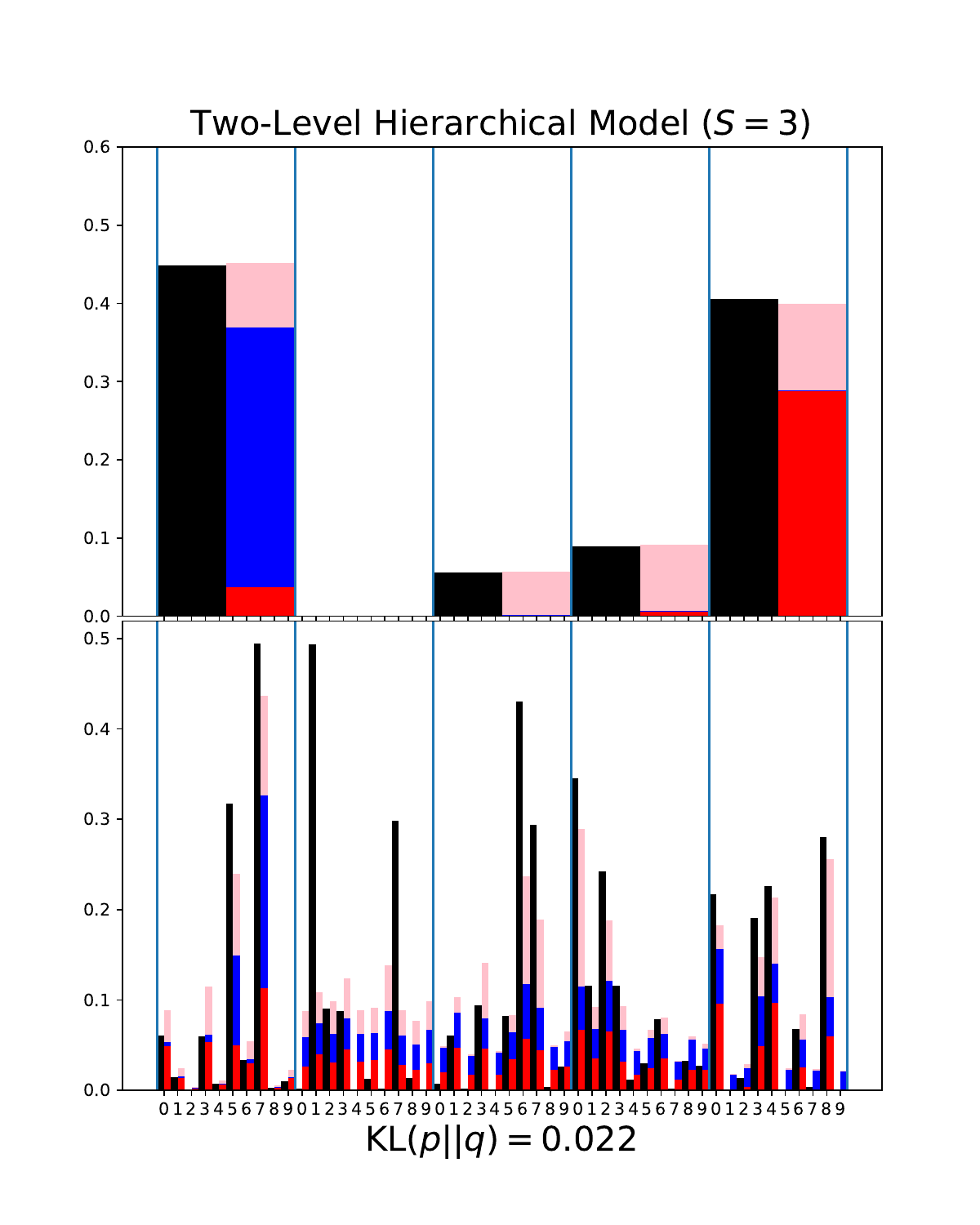}
    \caption{}
    \label{fig:my_label_3}
\end{subfigure}~
\begin{subfigure}{0.33\textwidth}
    \includegraphics[width=\textwidth]{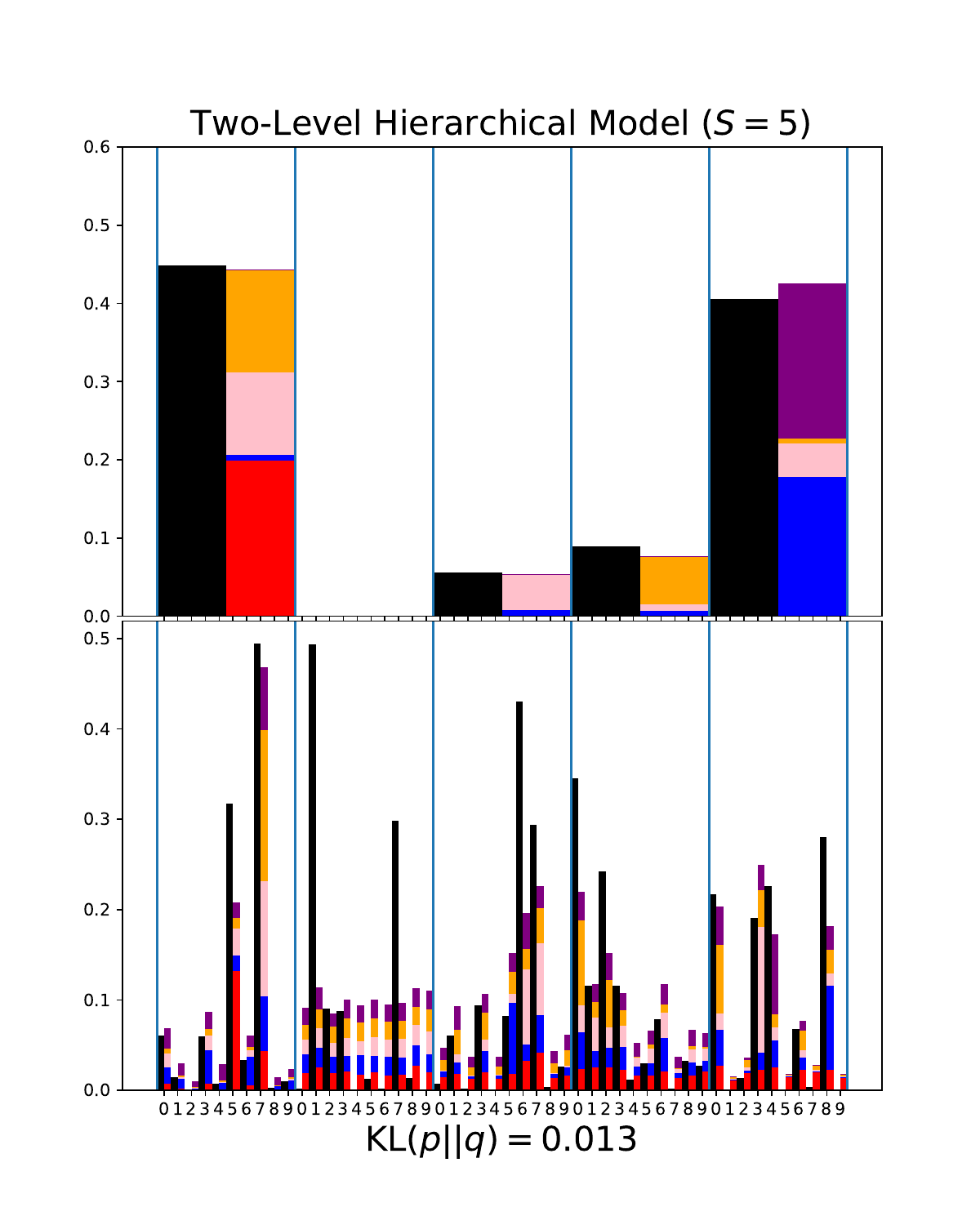}
    \caption{}
    \label{fig:my_label_4}
\end{subfigure}
\caption{Approximations using (a) $S=1$, (b) $S=3$ and (c) $S=5$ components of a two-level hierarchical target distribution with $n_1 = 5$ and $n_2=10$. The target is plotted in black. Each bin in the lower plot, delimited by the blue vertical lines, contains a CPD, conditioned on $z_1$ in the upper plot. The colors of the approximations represent the different mixture components, and their probabilities, scaled by $1/S$, are stacked. 
The approximations were trained using $K = \lfloor 20 / S\rfloor$ importance samples, and $S=5$ achieved the smallest KL divergence from the target distribution, printed below the plots. The components clearly cooperate in (b-c) as they explore complementary parts of the solution space.
}
\label{fig:bar_plot}
\end{figure}

\begin{wrapfigure}[17]{R}{0.35\textwidth}
    \centering
\includegraphics[width=0.35\textwidth]{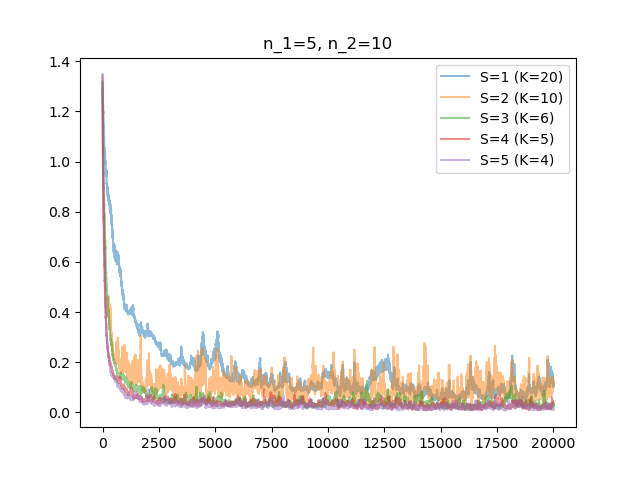}
    \caption{KL divergences from the target visualized in Fig. \ref{fig:bar_plot} and the approximations with $S=1,...,5$ mixture components. On the $x$-axis are the number of training iterations.}
    \label{fig:kl_curves}
\end{wrapfigure}

Using a two-leveled hierarchical model of categorical distributions as the target distribution,
$p(z_1,z_2)=p(z_{2}|z_{1})p(z_1)$, we wish to minimize $\text{KL}\left(\frac{1}{S}\sum_{s=1}^S q_{\phi_s}(z_1,z_2)\big\Vert p(z_1,z_2)\right)$. The CPD
 $p(z_{2}|z_{1})$ is a categorical distribution with $n_2$ categories, conditioned on the sampled category in the previous level, $z_{1}$, and $p(z_1)$ is a categorical distribution with $n_1$ categories. Similarly, the $s$-th component in the mixture approximation, $q_{\phi_s}(z_1,z_2) = q_{\phi_s}(z_2|z_1)q_{\phi_s}(z_1)$, is a two-level hierarchical model with learnable probabilities, $\phi_s$. The parameters of $p$ are drawn from a Dirichlet distribution with all concentration parameters equal to $0.5$, the approximations are trained using the VIMCO estimator derived in Sec. \ref{sec:vimco}, and the learning rates are chosen based on a grid search on a different target distribution. All $\phi_s$ are initialized uniformly over the categories.

 In Fig. \ref{fig:bar_plot}, three learned approximations are shown when $n_1=5$ and $n_2=10$, along with the corresponding KL divergences from the target to the approximation. For $S>1$, the components have spread out, exploring complementary parts of the solution space. Note that the $p(z_1=2)$ has a negligible probability, resulting in approximations that do not capture $p(z_2|z_1=2)$.

 We include the curves of the KL divergences over the training iterations when $n_1=5$ and $n_2=10$ in Fig. \ref{fig:kl_curves}, and, in Appendix \ref{app:twolevelmodel}, we show the KL curves for other choices of $K$, $n_1$ and $n_2$. In all cases where $p(z_1)$ has multiple categories with non-negligible probabilities, (relating to multimodality in the tree-topology posterior) mixtures converge with fewer iterations and to smaller KL divergences.

\subsection{Posterior Approximations using Real Data}
\label{sec:real data}
We performed experiments on eight datasets \citep{Hedges1990-eu, Garey1996-ti, Yang2003-eg, Henk2003-dn, Lakner2008-ks, Zhang2001-hs, Yoder2004-ut, Rossman2001-ph} which we will refer to as DS1-8. These are popular datasets for evaluating Bayesian phylogenetics methods, and, as in \cite{zhang2019variational, zhang2020improved, moretti2021variational, koptagelvaiphy, zhang2022variational, zhang2023learnable}, we focus on learning the approximations of branch-length and tree-topology distributions. Following the referenced works, we assume the exponential branch-length prior has rate 10 and a uniform prior over all unrooted trees (see Sec. \ref{sec:background} for details about the generative model). The substitution model is the Jukes-Cantor 69 model \citep{Jukes1969-gq}, and the candidate trees, $\mathcal{T}$, are gathered from ten replicates of~10000 ultrafast maximum likelihood bootstrap
trees \citep{Minh2013-mu}. The implementation is based on the code provided by \cite{zhang2022variational}, and we trained all VBPI models during~400,000 iterations, using the same hyperparameter settings as \cite{zhang2019variational, zhang2020improved}. Based on the study in \cite{Zhang2022-kt}, we let $K=10$ during training.

\begin{figure}[]
    \centering
    \begin{subfigure}{0.3\textwidth}
    \centering
    \includegraphics[width=\textwidth]{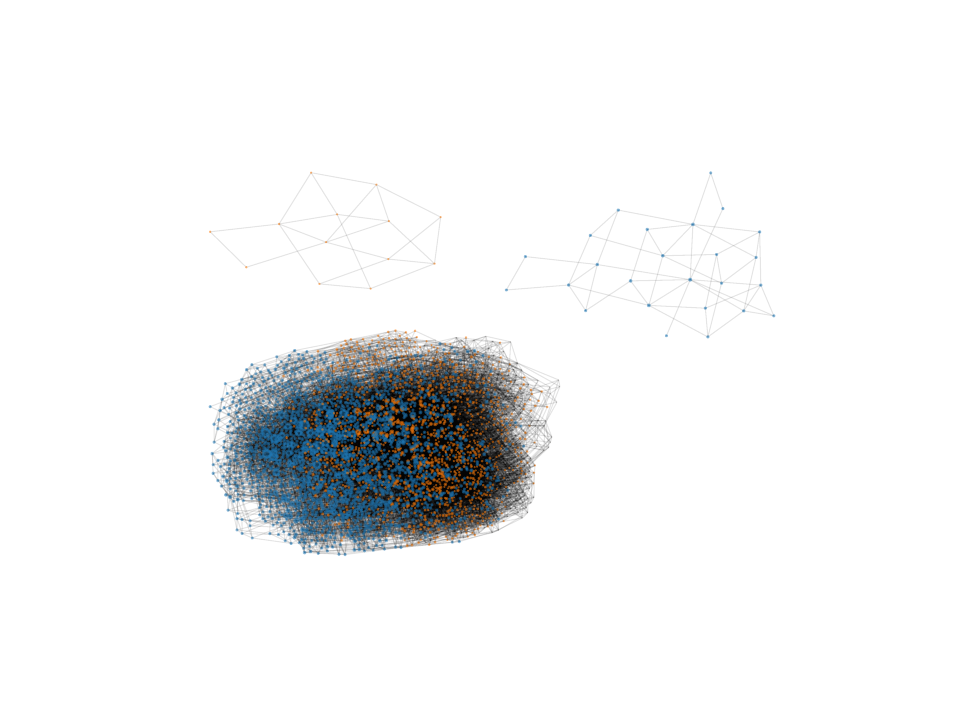}
    \end{subfigure}
    ~
    \begin{subfigure}{0.3\textwidth}
    \centering
    \includegraphics[width=\textwidth]{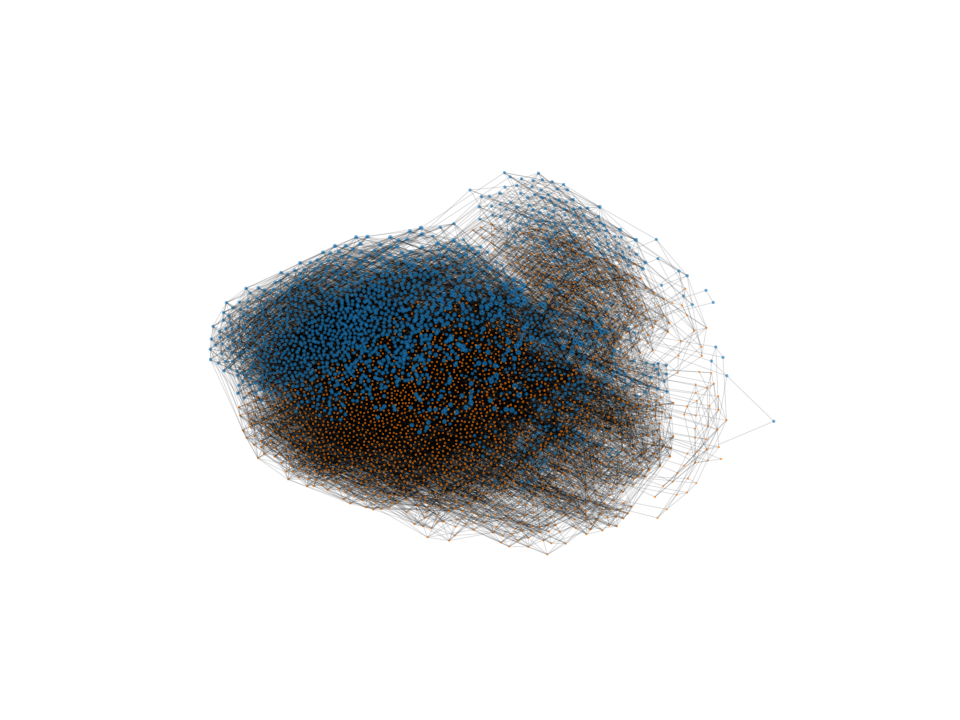}
    \end{subfigure}
    ~
    \begin{subfigure}{0.3\textwidth}
    \centering
    \includegraphics[width=\textwidth]{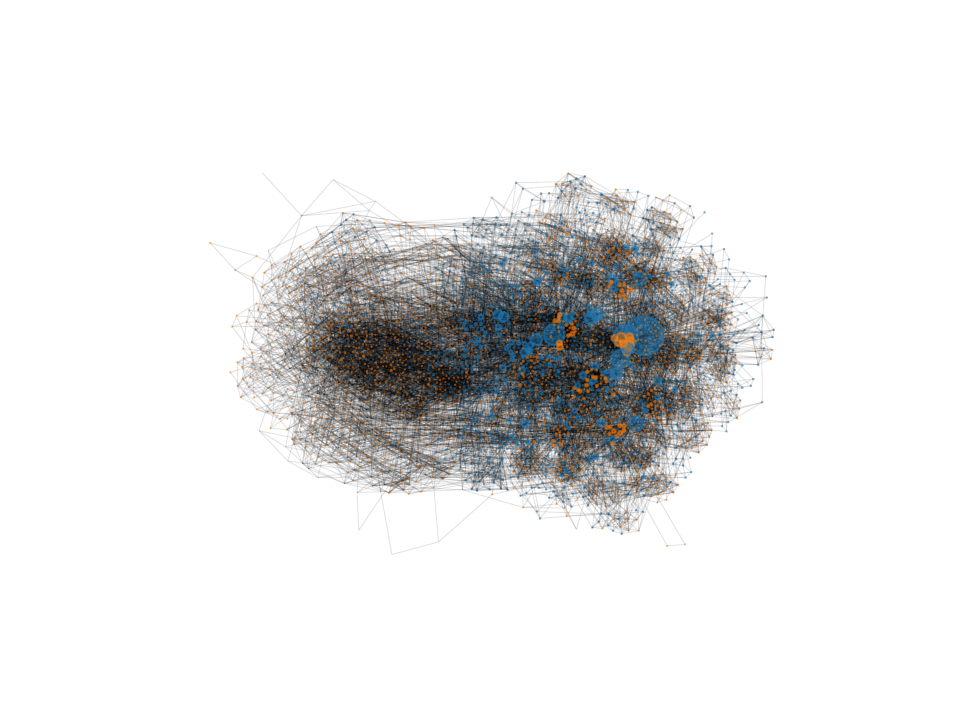}
    \end{subfigure}
    \begin{subfigure}{0.3\textwidth}
    \centering
    \includegraphics[width=\textwidth]{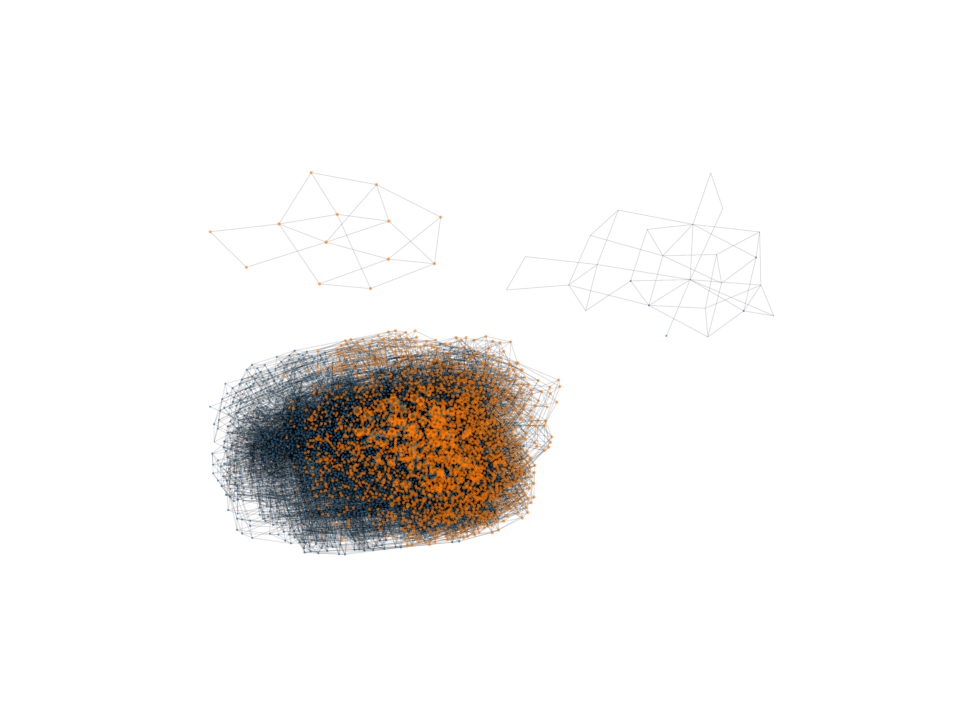}
    \caption{DS5}
    \end{subfigure}
    ~
    \begin{subfigure}{0.3\textwidth}
    \centering
    \includegraphics[width=\textwidth]{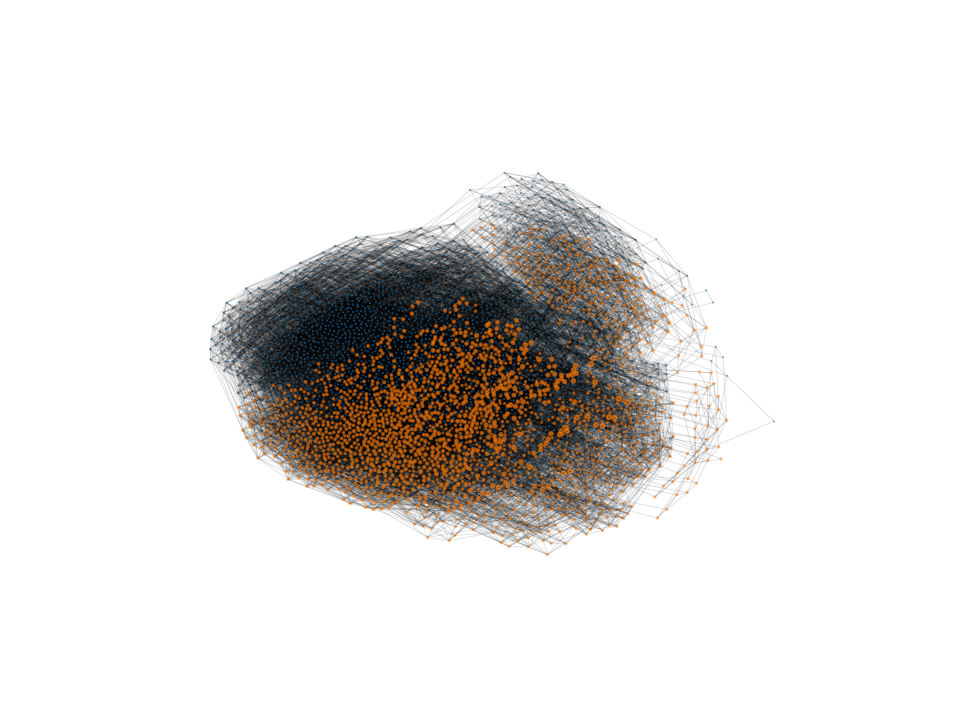}
    \caption{DS6}
    \end{subfigure}
    ~
    \begin{subfigure}{0.3\textwidth}
    \centering
    \includegraphics[width=\textwidth]{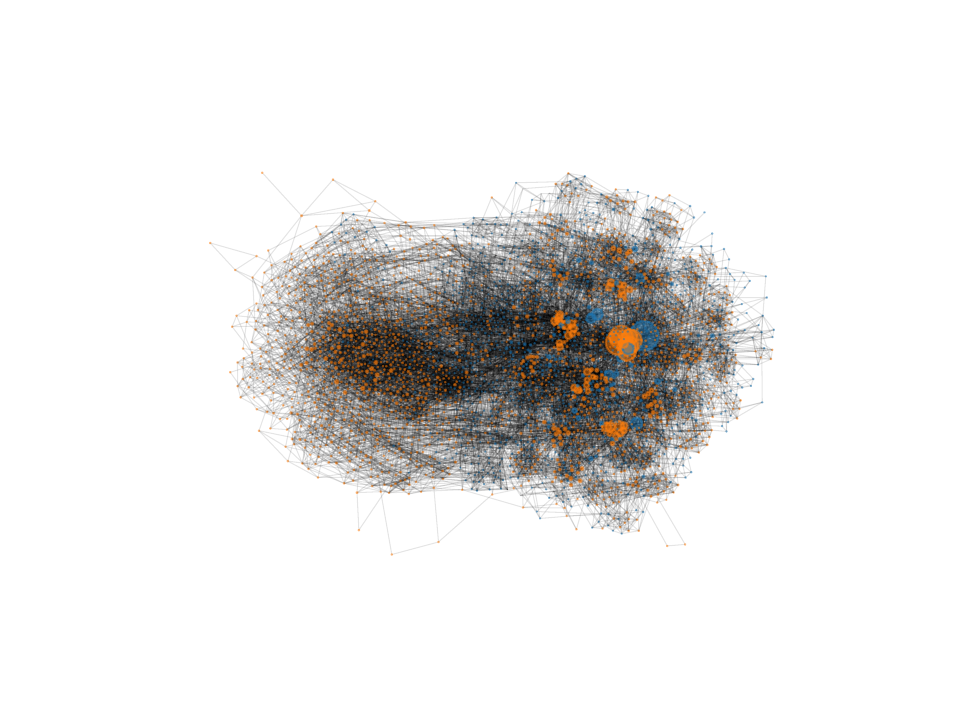}
    \caption{DS8}
    \end{subfigure}
    \caption{Visualization of a uniformly weighted $S=2$-component mixture of SBNs on (a) DS5, (b) DS6 and (c) DS8, where each node corresponds to a unique tree-topology. The upper row shows the distribution of five million sampled tree topologies from the first component, where a node, $\tau$, is colored blue if $q_{\phi_1}(\tau)>q_{\phi_2}(\tau)$, or orange otherwise. Vice versa for the lower row. The size of a node is determined by its sampling frequency, which is why nodes with low frequency appear black. The components clearly spread out, exploring different parts of the space.}
\label{fig:topology_visualization}
\end{figure}
\subsubsection{Visualizing the Explorative Behaviour of Mixtures of SBNs}
To graphically confirm the power of employing mixtures of SBNs, we in Fig. \ref{fig:topology_visualization} visualize representations of the learned tree-topology posterior approximations for a subset of the different real datasets. The subset was selected based on the datasets where the explorative behavior of the approximations is most clear. The representations for the other datasets, along with more implementation details, are included in the Appendix \ref{app:visualization}. 
 The SBNs correspond to VBPI-Mixtures without NFs. The components (upper vs. lower row) have jointly explored the space, partly specializing on disjoint sets of tree-topologies, verifying that the MISELBO objective promotes coordinated exploration of the discrete latent space, as discussed in Sec. \ref{sec:vbpi-mixtures}.

\subsubsection{Quantitative Evaluation of the Approximations}
\label{sec:quant_exp}
We quantitatively evaluate the approximations by, first, computing their KL divergences to the true posterior, and by, secondly, benchmarking their marginal log-likelihood estimates.
The results are averaged over five independently trained models with different parameter initializations. All MrBayes \citep{huelsenbeck2001mrbayes} results were produced using ten million long MCMC runs with four chains, sampling every 100 iterations. Our methods are denoted Mix$_{S}$, Mix$_{\text{NF}, S}$, representing VBPI-Mixtures with PSPs or NFs, respectively.
Mixtures that employ NFs share flow models, as described in \cite{kviman2023cooperation}.

\paragraph{Statistical distances to the tree-topology posterior}
\begin{table}[]
    \centering
    \caption{Illustrating the impact of VBPI-Mixtures in terms of KL$(p(\tau|X)\Vert 
 q_\phi(\tau))$. Lower is better.}
    
    \begin{tabular}{l|cccccccc}
 & DS1 & DS2 & DS3 & DS4 & DS5 & DS6 & DS7 & DS8 \\\hline
VBPI-NF & 0.0726 & 0.0110 &0.0540 &	0.2093 & 2.2117 & 1.2842 &	0.2544 & 0.6018 \\
Mix\textsubscript{NF,$S=2$} & 0.0631 &0.0059	&0.0475	&0.0965	&2.0337	&1.0883	&0.1183	&0.5199 \\
Mix\textsubscript{NF, $S=3$} & \textbf{0.0598}	&\textbf{0.0051}	&\textbf{0.0377}	&\textbf{0.0769}	&\textbf{1.9526	}&\textbf{1.0461	}&\textbf{0.0847}	&\textbf{0.4567} \\
\end{tabular}
    \label{tab:tv_tree}
\end{table}
Here, we compare statistical distances to the tree-topology posterior obtained from MrBayes as described above. 
In Table \ref{tab:tv_tree}, the KL divergence from the posterior to the approximations, i.e., KL$(p(\tau|X)\Vert 
 q_\phi(\tau))$, is computed, where $q_\phi$ represents a mixture of SBNs, or a single SBN, from VBPI-NF. Lower is better, and, notably, VBPI-Mixtures consistently produce KL divergences across all datasets that monotonically decrease with $S$. 

\begin{table}[]
    \centering
    \caption{Marginal log-likelihood estimates on DS1-8. All VBPI methods use $1000$ importance samples, and the results are averaged over~100 runs and three independently trained models. 
    Following \cite{zhang2019variational, zhang2020improved, zhang2022variational, zhang2023learnable}, we bold font the results with the lowest standard deviations (shown in parentheses). Details in Sec. \ref{sec:quant_exp}. Mixtures monotonically improve with $S$.}
    {\tiny
    \tabcolsep1.1pt
    \begin{tabularx}{\linewidth}{@{}X|XXXXXXXX@{}}
\hline
Data & DS1 & DS2 & DS3 & DS4 & DS5 & DS6 & DS7 & DS8 \\\hline
\# Taxa & 27 & 29 & 36 & 41 & 50 & 50 & 59 & 64 \\
\# Sites & 1949 & 2520 & 1812 & 1137 & 378 & 1133 & 1824 & 1008 \\\hline
\multicolumn{9}{c}{\textbf{VBPI with Mixtures}}\\
\hline
VBPI & -7108.50 (0.23)& -26367.70 (0.09)& -33735.10 (0.14)& -13330.03 (0.23)& -8214.80 (0.50)& -6724.59 (0.53)& -37332.12 (0.45)& -8652.39 (0.71) \\
Mix\textsubscript{$S=2$} & -7108.44 (0.12)& -26367.71 (0.06)& -33735.10 (0.07)& -13330.00 (0.17)& -8214.75 (0.36)& -6724.54 (0.31)& -37332.04 (0.24)& -8651.68 (0.49) \\
Mix\textsubscript{$S=3$} & -7108.42 (0.11) &	-26367.71 (0.04) &	-33735.10 (0.06) &	-13329.97 (0.17) &	-8214.73 (0.26) &	-6724.51 (0.28) &	-37332.03 (0.18) &	-8650.83 (0.46) \\
\hline
\multicolumn{9}{c}{\textbf{VBPI with NFs and Mixtures}}\\
\hline
VBPI-NF & -7108.42 (0.15)& -26367.72 (0.06)& -33735.10 (0.07)& -13330.00 (0.23)& -8214.70 (0.47)& -6724.50 (0.45)& -37332.01 (0.27)& -8650.68 (0.46) \\
Mix\textsubscript{NF,$S=2$} & -7108.40 (0.10)& -26367.71 (0.04)& -33735.10 (0.05)& -13329.95 (0.15)& -8214.62 (0.26)& -6724.44 (0.32)& -37331.96 (0.19)& -8650.56 (0.33) \\
Mix\textsubscript{NF,$S=3$} & \textbf{-7108.40 (0.06)}& \textbf{-26367.70 (0.03)}& \textbf{-33735.09 (0.04)}& \textbf{-13329.94 (0.11)}& \textbf{-8214.56 (0.22)}& \textbf{-6724.40 (0.23)}& \textbf{-37331.96 (0.15)}& \textbf{-8650.54 (0.30)} \\
\hline
\multicolumn{9}{c}{\textbf{MCMC and VBPI with GNNs (scores from \cite{zhang2019variational} and \cite{zhang2023learnable})}}\\
\hline
MrBayes\textsubscript{ss} & -7108.42 (0.18) & -26367.57 (0.48) & -33735.44 (0.50) & -13330.06 (0.54) & -8214.51 (0.28) & -6724.07 (0.86) & -37332.76 (2.42) & -8649.88 (1.75) \\
GGNN & -7108.40 (0.19) & -26367.73 (0.10) & -33735.11 (0.09) & -13329.95 (0.19) & -8214.67 (0.36) & -6724.38 (0.42) & -37332.03 (0.30) & -8650.68 (0.48) \\
EDGE & -7108.41 (0.14) & -26367.73 (0.07) & -33735.12 (0.09) & -13329.94 (0.19) & -8214.64 (0.38) & -6724.37 (0.40) & -37332.04 (0.26) & -8650.65 (0.45)
\end{tabularx}
    }
    \label{tab:ml_elbo}
    \vspace*{-1.01\baselineskip}
\end{table}

\paragraph{Marginal log-likelihood estimates}
In terms of marginal log-likelihood estimates, we benchmark our methods against the existing VBPI algorithms: VBPI with PSP parameterization \citep{zhang2019variational}, VBPI-NF \citep{zhang2020improved} with ten RealNVPs \citep{dinh2016density}, and VBPI-GNN (\cite{zhang2023learnable}; EDGE and GGNN). Additionally, we compare our results with the stepping-stone (SS; \cite{xie2011improving}) method applied to MrBayes. The results are given in Table \ref{tab:ml_elbo}. 
Following \cite{zhang2019variational, zhang2020improved, zhang2022variational, zhang2023learnable}, we bold font the results with the lowest standard deviations. Rewarding low-variance estimates is motivated, as they imply, for instance, more reliable Bayesian model selections for downstream tasks. 
Increasing the number of mixture components results in significant improvements in terms of lower standard deviations (on all datasets) and higher mean log-likelihood scores (especially apparent on the more complex datasets, e.g. DS5-8).

\section{Conclusion}
We introduced VBPI-Mixtures, a novel algorithm that increases the flexibility of the phylogenetic posterior approximation by utilizing recent advances in mixtures for black-box VI. We showed that mixtures of SBNs can approximate distributions that a single SBN cannot, making a persuasive case for VBPI-Mixtures. Experimentally, we achieved state-of-the-art results in terms of marginal log-likelihood estimation and produced more accurate approximations of the tree-topology posterior. 

\section*{Acknowledgments}
This project was made possible through funding from the
Swedish Foundation for Strategic Research grants BD15-
0043 and ID19-0052, from the Swedish Research Council grant 2018-05417\_VR, and was supported by the Wallenberg AI, Autonomous Systems and Software Program (WASP) funded by the Knut and Alice Wallenberg Foundation. The computations and data handling were enabled by resources provided by the Swedish
National Infrastructure for Computing (SNIC), partially
funded by the Swedish Research Council through grant
agreement no. 2018-05973.

\bibliography{iclr2024_conference}

\begin{thebibliography}{38}
\providecommand{\natexlab}[1]{#1}
\providecommand{\url}[1]{\texttt{#1}}
\expandafter\ifx\csname urlstyle\endcsname\relax
  \providecommand{\doi}[1]{doi: #1}\else
  \providecommand{\doi}{doi: \begingroup \urlstyle{rm}\Url}\fi

\bibitem[Bouchard-C{\^o}t{\'e} et~al.(2012)Bouchard-C{\^o}t{\'e}, Sankararaman, and Jordan]{bouchard2012phylogenetic}
Alexandre Bouchard-C{\^o}t{\'e}, Sriram Sankararaman, and Michael~I Jordan.
\newblock Phylogenetic inference via sequential monte carlo.
\newblock \emph{Systematic biology}, 61\penalty0 (4):\penalty0 579--593, 2012.

\bibitem[Burda et~al.(2016)Burda, Grosse, and Salakhutdinov]{Burda2015-lr}
Yuri Burda, Roger Grosse, and Ruslan Salakhutdinov.
\newblock Importance weighted autoencoders.
\newblock In \emph{ICLR}. PMLR, 2016.

\bibitem[Dinh et~al.(2016)Dinh, Sohl-Dickstein, and Bengio]{dinh2016density}
Laurent Dinh, Jascha Sohl-Dickstein, and Samy Bengio.
\newblock Density estimation using real nvp.
\newblock \emph{arXiv preprint arXiv:1605.08803}, 2016.

\bibitem[Felsenstein(2003)]{Felsenstein2003-ts}
Joseph Felsenstein.
\newblock \emph{Inferring Phylogenies}.
\newblock Sinauer, October 2003.

\bibitem[Garey et~al.(1996)Garey, Near, Nonnemacher, and {others}]{Garey1996-ti}
J~R Garey, T~J Near, M~R Nonnemacher, and {others}.
\newblock Molecular evidence for acanthocephala as a subtaxon of rotifera.
\newblock \emph{Journal of Molecular}, 1996.

\bibitem[Hagberg et~al.(2008)Hagberg, Swart, and S~Chult]{hagberg2008exploring}
Aric Hagberg, Pieter Swart, and Daniel S~Chult.
\newblock Exploring network structure, dynamics, and function using networkx.
\newblock Technical report, Los Alamos National Lab. (LANL), Los Alamos, NM (United States), 2008.

\bibitem[Hedges et~al.(1990)Hedges, Moberg, and {others}]{Hedges1990-eu}
S~B Hedges, K~D Moberg, and {others}.
\newblock Tetrapod phylogeny inferred from {18S} and {28S} ribosomal {RNA} sequences and a review of the evidence for amniote relationships.
\newblock \emph{Mol. Biol.}, 1990.

\bibitem[Henk et~al.(2003)Henk, Weir, and Blackwell]{Henk2003-dn}
Daniel~A Henk, Alex Weir, and Meredith Blackwell.
\newblock Laboulbeniopsis termitarius, an ectoparasite of termites newly recognized as a member of the laboulbeniomycetes.
\newblock \emph{Mycologia}, 95\penalty0 (4):\penalty0 561--564, 2003.

\bibitem[H{\"o}hna et~al.(2016)H{\"o}hna, Landis, Heath, Boussau, Lartillot, Moore, Huelsenbeck, and Ronquist]{hohna2016revbayes}
Sebastian H{\"o}hna, Michael~J Landis, Tracy~A Heath, Bastien Boussau, Nicolas Lartillot, Brian~R Moore, John~P Huelsenbeck, and Fredrik Ronquist.
\newblock Revbayes: Bayesian phylogenetic inference using graphical models and an interactive model-specification language.
\newblock \emph{Systematic biology}, 65\penalty0 (4):\penalty0 726--736, 2016.

\bibitem[Huelsenbeck \& Ronquist(2001)Huelsenbeck and Ronquist]{huelsenbeck2001mrbayes}
John~P Huelsenbeck and Fredrik Ronquist.
\newblock Mrbayes: Bayesian inference of phylogenetic trees.
\newblock \emph{Bioinformatics}, 17\penalty0 (8):\penalty0 754--755, 2001.

\bibitem[Jukes et~al.(1969)Jukes, Cantor, and {Others}]{Jukes1969-gq}
Thomas~H Jukes, Charles~R Cantor, and {Others}.
\newblock Evolution of protein molecules.
\newblock \emph{Mammalian protein metabolism}, 3:\penalty0 21--132, 1969.

\bibitem[Kipf \& Welling(2017)Kipf and Welling]{kipf2016semi}
Thomas~N Kipf and Max Welling.
\newblock Semi-supervised classification with graph convolutional networks.
\newblock \emph{ICLR}, 2017.

\bibitem[Koptagel et~al.(2022)Koptagel, Kviman, Melin, Safinianaini, and Lagergren]{koptagelvaiphy}
Hazal Koptagel, Oskar Kviman, Harald Melin, Negar Safinianaini, and Jens Lagergren.
\newblock Vaiphy: a variational inference based algorithm for phylogeny.
\newblock In \emph{Advances in Neural Information Processing Systems}, 2022.

\bibitem[Kviman et~al.(2022)Kviman, Melin, Koptagel, Elvira, and Lagergren]{kviman2022multiple}
Oskar Kviman, Harald Melin, Hazal Koptagel, Victor Elvira, and Jens Lagergren.
\newblock Multiple importance sampling elbo and deep ensembles of variational approximations.
\newblock In \emph{International Conference on Artificial Intelligence and Statistics}, pp.\  10687--10702. PMLR, 2022.

\bibitem[Kviman et~al.(2023)Kviman, Mol{\'e}n, Hotti, Kurt, Elvira, and Lagergren]{kviman2023cooperation}
Oskar Kviman, Ricky Mol{\'e}n, Alexandra Hotti, Semih Kurt, V{\i}ctor Elvira, and Jens Lagergren.
\newblock Cooperation in the latent space: The benefits of adding mixture components in variational autoencoders.
\newblock In \emph{International Conference on Machine Learning}, pp.\  18008--18022. PMLR, 2023.

\bibitem[Lakner et~al.(2008)Lakner, van~der Mark, Huelsenbeck, Larget, and Ronquist]{Lakner2008-ks}
Clemens Lakner, Paul van~der Mark, John~P Huelsenbeck, Bret Larget, and Fredrik Ronquist.
\newblock Efficiency of markov chain monte carlo tree proposals in bayesian phylogenetics.
\newblock \emph{Syst. Biol.}, 57\penalty0 (1):\penalty0 86--103, February 2008.

\bibitem[Minh et~al.(2013)Minh, Nguyen, and von Haeseler]{Minh2013-mu}
Bui~Quang Minh, Minh Anh~Thi Nguyen, and Arndt von Haeseler.
\newblock Ultrafast approximation for phylogenetic bootstrap.
\newblock \emph{Mol. Biol. Evol.}, 30\penalty0 (5):\penalty0 1188--1195, May 2013.

\bibitem[Mnih \& Rezende(2016)Mnih and Rezende]{Mnih2016-fp}
Andriy Mnih and Danilo Rezende.
\newblock Variational inference for monte carlo objectives.
\newblock In Maria~Florina Balcan and Kilian~Q Weinberger (eds.), \emph{Proceedings of The 33rd International Conference on Machine Learning}, volume~48 of \emph{Proceedings of Machine Learning Research}, pp.\  2188--2196, New York, New York, USA, 2016. PMLR.

\bibitem[Moretti et~al.(2021)Moretti, Zhang, Naesseth, Venner, Blei, and Pe’er]{moretti2021variational}
Antonio~Khalil Moretti, Liyi Zhang, Christian~A Naesseth, Hadiah Venner, David Blei, and Itsik Pe’er.
\newblock Variational combinatorial sequential monte carlo methods for bayesian phylogenetic inference.
\newblock In \emph{Uncertainty in Artificial Intelligence}, pp.\  971--981. PMLR, 2021.

\bibitem[Morningstar et~al.(2021)Morningstar, Vikram, Ham, Gallagher, and Dillon]{morningstar2021automatic}
Warren Morningstar, Sharad Vikram, Cusuh Ham, Andrew Gallagher, and Joshua Dillon.
\newblock Automatic differentiation variational inference with mixtures.
\newblock In \emph{International Conference on Artificial Intelligence and Statistics}, pp.\  3250--3258. PMLR, 2021.

\bibitem[Nalisnick et~al.(2016)Nalisnick, Hertel, and Smyth]{nalisnick2016approximate}
Eric Nalisnick, Lars Hertel, and Padhraic Smyth.
\newblock Approximate inference for deep latent gaussian mixtures.
\newblock In \emph{NIPS Workshop on Bayesian Deep Learning}, volume~2, pp.\  131, 2016.

\bibitem[Ranganath et~al.(2014)Ranganath, Gerrish, and Blei]{ranganath2014black}
Rajesh Ranganath, Sean Gerrish, and David Blei.
\newblock Black box variational inference.
\newblock In \emph{Artificial intelligence and statistics}, pp.\  814--822. PMLR, 2014.

\bibitem[Rezende \& Mohamed(2015)Rezende and Mohamed]{rezende2015variational}
Danilo Rezende and Shakir Mohamed.
\newblock Variational inference with normalizing flows.
\newblock In \emph{International conference on machine learning}, pp.\  1530--1538. PMLR, 2015.

\bibitem[Rossman et~al.(2001)Rossman, McKemy, Pardo-Schultheiss, and Schroers]{Rossman2001-ph}
Amy~Y Rossman, John~M McKemy, Rebecca~A Pardo-Schultheiss, and Hans-Josef Schroers.
\newblock Molecular studies of the bionectriaceae using large subunit {rDNA} sequences.
\newblock \emph{Mycologia}, 93\penalty0 (1):\penalty0 100--110, January 2001.

\bibitem[Ruiz et~al.(2016)Ruiz, Titsias, and Blei]{ruiz2016overdispersed}
Francisco~JR Ruiz, Michalis~K Titsias, and David~M Blei.
\newblock Overdispersed black-box variational inference.
\newblock \emph{Uncertainty in Artificial Intelligence}, 2016.

\bibitem[Wang et~al.(2015)Wang, Bouchard-C{\^o}t{\'e}, and Doucet]{wang2015bayesian}
Liangliang Wang, Alexandre Bouchard-C{\^o}t{\'e}, and Arnaud Doucet.
\newblock Bayesian phylogenetic inference using a combinatorial sequential monte carlo method.
\newblock \emph{Journal of the American Statistical Association}, 110\penalty0 (512):\penalty0 1362--1374, 2015.

\bibitem[Wang et~al.(2020)Wang, Wang, and Bouchard-C{\^o}t{\'e}]{wang2020annealed}
Liangliang Wang, Shijia Wang, and Alexandre Bouchard-C{\^o}t{\'e}.
\newblock An annealed sequential monte carlo method for bayesian phylogenetics.
\newblock \emph{Systematic biology}, 69\penalty0 (1):\penalty0 155--183, 2020.

\bibitem[Whidden \& Matsen~IV(2015)Whidden and Matsen~IV]{whidden2015quantifying}
Chris Whidden and Frederick~A Matsen~IV.
\newblock Quantifying mcmc exploration of phylogenetic tree space.
\newblock \emph{Systematic biology}, 64\penalty0 (3):\penalty0 472--491, 2015.

\bibitem[Xie et~al.(2011)Xie, Lewis, Fan, Kuo, and Chen]{xie2011improving}
Wangang Xie, Paul~O Lewis, Yu~Fan, Lynn Kuo, and Ming-Hui Chen.
\newblock Improving marginal likelihood estimation for bayesian phylogenetic model selection.
\newblock \emph{Systematic biology}, 60\penalty0 (2):\penalty0 150--160, 2011.

\bibitem[Yang \& Yoder(2003)Yang and Yoder]{Yang2003-eg}
Z~Yang and A~D Yoder.
\newblock Comparison of likelihood and bayesian methods for estimating divergence times using multiple gene loci and calibration points, with application to a radiation of cute-looking mouse lemur species.
\newblock \emph{Syst. Biol.}, 2003.

\bibitem[Yoder \& Yang(2004)Yoder and Yang]{Yoder2004-ut}
Anne~D Yoder and Ziheng Yang.
\newblock Divergence dates for malagasy lemurs estimated from multiple gene loci: geological and evolutionary context.
\newblock \emph{Mol. Ecol.}, 13\penalty0 (4):\penalty0 757--773, April 2004.

\bibitem[Zhang(2020)]{zhang2020improved}
Cheng Zhang.
\newblock Improved variational bayesian phylogenetic inference with normalizing flows.
\newblock \emph{Advances in neural information processing systems}, 33:\penalty0 18760--18771, 2020.

\bibitem[Zhang(2023)]{zhang2023learnable}
Cheng Zhang.
\newblock Learnable topological features for phylogenetic inference via graph neural networks.
\newblock \emph{ICLR}, 2023.

\bibitem[Zhang \& Matsen(2022)Zhang and Matsen]{Zhang2022-kt}
Cheng Zhang and Frederick~A Matsen, IV.
\newblock A variational approach to bayesian phylogenetic inference.
\newblock April 2022.

\bibitem[Zhang \& Matsen~IV(2018)Zhang and Matsen~IV]{zhang2018generalizing}
Cheng Zhang and Frederick~A Matsen~IV.
\newblock Generalizing tree probability estimation via bayesian networks.
\newblock \emph{Advances in neural information processing systems}, 31, 2018.

\bibitem[Zhang \& Matsen~IV(2019)Zhang and Matsen~IV]{zhang2019variational}
Cheng Zhang and Frederick~A Matsen~IV.
\newblock Variational bayesian phylogenetic inference.
\newblock In \emph{ICLR}, 2019.

\bibitem[Zhang \& Matsen~IV(2022)Zhang and Matsen~IV]{zhang2022variational}
Cheng Zhang and Frederick~A Matsen~IV.
\newblock A variational approach to bayesian phylogenetic inference.
\newblock \emph{arXiv preprint arXiv:2204.07747}, 2022.

\bibitem[Zhang \& Blackwell(2001)Zhang and Blackwell]{Zhang2001-hs}
N~Zhang and M~Blackwell.
\newblock Molecular phylogeny of dogwood anthracnose fungus (discula destructiva) and the diaporthales.
\newblock \emph{Mycologia}, 2001.

\end{thebibliography}
\bibliographystyle{iclr2024_conference}

\appendix
\section{A Brief Introduction to Bayesian Phylogenetic Inference for Machine Learning Researchers}
\label{sec:intro_to_phylo}
This introduction aims to briefly explain the basic concepts required to understand the generative model provided in the main text.

Phylogenetic trees capture evolutionary relationships among species and provides valuable insights into life's evolutionary history. Within this domain, phylogenies are often depicted as bifurcating tree graphs, where nodes represent common ancestors, and branches (edges) signify evolutionary events and genetic distances between species. This framework enables an understanding of species relatedness, ancestry, and the evolutionary processes governing life's diversity.

Bayesian phylogenetic inference builds upon this framework by applying Bayesian statistical methods to infer the evolutionary history. It allows for a probabilistic approach to model uncertainty and variation, considering prior beliefs about evolutionary parameters and updating these beliefs as new data is incorporated. It is common to use DNA or protein sequences as data since it describes different attributes of the species, and the edges represent a mutation between species. Through sampling from a posterior distribution and utilizing tools like Markov Chain Monte Carlo (MCMC) methods or variational inference, Bayesian phylogenetics offers a robust and nuanced view of evolutionary relationships, integrating multiple sources of information and providing a rigorous statistical foundation for evolutionary hypotheses.

Mainly, two latent variables are regarded as important in Bayesian phylogenetic inference. First, the tree topology, $\tau$, a binary tree with the observations assigned to its leaves. The tree-topology space grows as $(2n - 3)!!$
 for rooted and $(2n - 5)!!$
 for unrooted trees, where $n$
 is the number of leaves (observations/taxa). Furthermore, each edge, $e$, of the topology is associated with a positive continuous variable, the branch lengths, $b(e)$. The Cartesian product of discrete and continuous spaces makes inference in phylogenetics a challenging task.

\section{Conflicting Tree Topologies}
 \label{sec:DNA-example}
Here we construct a realistic scenario where DNA sequences induce conflicting tree-topologies in the posterior that cannot be modelled by the vanilla SBN. On the other hand, they can be captured by VBPI-Mixtures.
 
 It is well-known that DNA data sometimes has conflicting signals. Here we construct a toy example to demonstrate how $\tau_1$ and $\tau_2$ can have higher posterior support than $\tau_3$ and $\tau_4$. For simplicity, we use the connection between a lower parsimony score and a higher likelihood when branch lengths are short (to appear in the appendix). First, consider that the leaves in Fig. \ref{fig:four trees} have DNA sequences with nucleotides at sites $i$ and $j$ specified in the table below (the first and second columns represent $i$ and $j$, respectively).

\begin{table}[ht!]
    \centering
    \caption{Nucleotide assignments to sites $i$ and $j$ in the observations.}
    {\tiny
    \tabcolsep1.1pt
    \begin{tabularx}{\linewidth}{@{}X|XXXXXX@{}}
Sites & (1) & (2) & (3) & (4) & (5) & (6)  \\\hline
$i$ & A & C & C & A & G & G \\
$j$ & C & C & A & G & G & A 
\end{tabularx}
    }
    \label{tab:realistic_dna_example}
\end{table}

We compute the parsimony scores for each clade (A, A', B, B') and the cost of joining two clades to form $\tau_1,...,\tau_4$. The cost of transitioning from one nucleotide (A, C, G, T) to another is 1.

In the first table below, we start by calculating all possible parsimony scores for each clade. Here X→Y denotes the parsimony score, Y, if the nucleotide at site $i$ or $j$ in the ancestral sequence in the root is X. Bolded is the best (lowest) score.

\begin{table}[ht!]
    \centering
    \caption{Parsimony scores for the four different subtrees}
    {\tiny
    \tabcolsep1.1pt
    \begin{tabularx}{\linewidth}{@{}X|XXXX@{}}
MP & $A=((1,2),3) $ & $B=((4,5), 6) $ & $A' = (1,(2,3)) $ & $B' = (4,(5,6)) $   \\\hline
$i$:th & A→2,\textbf{C→1},G→3,T→3 & A→2,C→3,\textbf{G→1},T→3 & \textbf{A→1},\textbf{C→1},G→2,T→2 & \textbf{A→1},C→2,\textbf{G→1},T→2 \\
$j$:th & \textbf{A→1},\textbf{C→1},G→2,T→2 & \textbf{A→1},C→3,\textbf{G→1},T→3 & A→2,\textbf{C→1},G→3,T→3 & A→2,C→3,\textbf{G→1},T→3
\end{tabularx}
    }
    \label{tab:realistic_dna_example_mp1}
\end{table}

\begin{table}[ht!]
    \centering
    \caption{Parsimony scores for all possible tree topologies}
    {\tiny
    \tabcolsep1.1pt
    \begin{tabularx}{\linewidth}{@{}X|XXXX@{}}
MP & $\tau_1 (A \wedge B)$ & $\tau_2 (A' \wedge  B') $ & $\tau_3 (A \wedge B') $ & $\tau_4 (A' \wedge B)$   \\\hline
$i$:th & A→4,\textbf{C→3},\textbf{G→3},T→4	 & \textbf{A→2},C→3,G→3,T→4 & \textbf{A→3},\textbf{C→3},\textbf{G→3},T→4 & \textbf{A→3},\textbf{C→3},\textbf{G→3},T→4 \\
$j$:th & \textbf{A→2},C→3,G→3,T→4 & A→4,\textbf{C→3},\textbf{G→3},T→4 & \textbf{A→3},\textbf{C→3},\textbf{G→3},T→4 & \textbf{A→3},\textbf{C→3},\textbf{G→3},T→4
\end{tabularx}
    }
    \label{tab:realistic_dna_example_mp2}
\end{table}
Now we can see that $\tau_1$ and $\tau_2$ give us a better parsimony score, demonstrating that the scenario exemplified in Fig. \ref{fig:four trees} can occur in biological data when there are conflicting signals.

We direct readers who want to learn more about parsimony scores to Chapter 1 in \cite{Felsenstein2003-ts}.
\section{Gradient Derivation}
\label{app:gradient_derivations}
Here we show the full derivations of the gradients w.r.t. $\phi_i$. For completeness, recall that 
\begin{equation}
    f(x, \mathcal{B}^k_s, \tau^k_s)=  \frac{p(\mathcal{B}^k_s, \tau^k_s, X)}{\frac{1}{S}\sum_{j=1}^S q_{\psi_j}(\mathcal{B}^k_s| \tau^k_s)q_{\phi_j}(\tau^k_s)},
\end{equation} 
and $\hat{L}^K_s = \log \frac{1}{K}\sum_{k=1}^K f(x, \mathcal{B}^k_s, \tau^k_s)$, where $\mathcal{B}^k_s, \tau^k_s$ are simulated from $ q_{\psi_s, \phi_s}(\mathcal{B}, \tau)$.

That is, we are interested in the gradient of Eq. (\ref{eq:miselbo}) w.r.t. the SBN parameters for one of the mixture components, say $i$,
\begin{equation}
\label{eq:grad_i_miselbo_appendix}
    \nabla_{\phi_i} \mathcal{L}(X;K, S) = \nabla_{\phi_i} \frac{1}{S}\sum_{s=1}^S\mathbb{E}_{q_{\psi_s,\phi_s}(\mathcal{B}, \tau)}\left[
    \hat{L}^K_s
    \right].
\end{equation}
There are two cases to take into account in the sum, either $i=s$ or $i\neq s$. Starting with $i=s$ and using the product rule,
\begin{align}
\nabla_{\phi_i}\frac{1}{S}\mathbb{E}_{\mathcal{B}_i^{1:K}, \tau_i^{1:K}\sim q_{\psi_i,\phi_i}(\mathcal{B}, \tau)}\left[ \hat{L}^K_i \right]
&= \nabla_{\phi_i}\frac{1}{S}\sum_{\tau_i^{1:K}}q_{\phi_i}(\tau_i^{1:K}) \mathbb{E}_{\mathcal{B}_i^{1:K}\sim q_{\psi_i}(\mathcal{B}|\tau_i^{1:K})}\Big[\hat{L}^K_i\Big]\\
&= \frac{1}{S}\sum_{\tau_i^{1:K}}\mathbb{E}_{\mathcal{B}_i^{1:K}\sim q_{\psi_i}(\mathcal{B}|\tau_i^{1:K})}\Big[ \nabla_{\phi_i} q_{\phi_i}(\tau_i^{1:K})\hat{L}^K_i\Big]\\\label{eq:q_grad}
&= \frac{1}{S}\sum_{\tau_i^{1:K}}\mathbb{E}_{q_{\psi_i}(\mathcal{B}|\tau_i^{1:K})}\Big[\hat{L}^K_i \nabla_{\phi_i} q_{\phi_i}(\tau_i^{1:K})+ q_{\phi_i}(\tau_i^{1:K}) \nabla_{\phi_i} \hat{L}^K_i \Big].
\end{align}
Recalling the identity that $\nabla_\phi g_\phi(z) = g_\phi(z)\nabla_\phi \log g_\phi(z)$, we start by rewriting the first term inside the expectation in Eq. (\ref{eq:q_grad}) as
\begin{align}
    \hat{L}^K_i \nabla_{\phi_i} q_{\phi_i}(\tau_i^{1:K}) &=  q_{\phi_i}(\tau_i^{1:K})\hat{L}^K_i\nabla_{\phi_i} \log q_{\phi_i}(\tau_i^{1:K})\\
    &= \label{eq:first_exp_term}q_{\phi_i}(\tau_i^{1:K})\hat{L}^K_i\sum_{k=1}^K\nabla_{\phi_i} \log q_{\phi_i}(\tau_i^{k}),
\end{align}
and then the second term
\begin{align}
   q_{\phi_i}(\tau_i^{1:K}) \nabla_{\phi_i} \hat{L}^K_i &=  q_{\phi_i}(\tau_i^{1:K})\nabla_{\phi_i} \log\frac{1}{K}\sum_{k=1}^K f(x, \mathcal{B}^k_i, \tau^k_i)\\
   &= q_{\phi_i}(\tau_i^{1:K}) \sum_{k=1}^K\tilde{w}^k_i\nabla_{\phi_i}\log f(x, \mathcal{B}^k_i, \tau^k_i)\\
   &=\label{eq:second_exp_term} -q_{\phi_i}(\tau_i^{1:K}) \sum_{k=1}^K\tilde{w}^k_i\nabla_{\phi_i}\log \frac{1}{S}\sum_{j=1}^S q_{\psi_j}(\mathcal{B}^k_i| \tau^k_i)q_{\phi_j}(\tau^k_i),
\end{align}
where $\tilde{w}^k_i = \frac{f(x, \mathcal{B}^k_i, \tau^k_i)}{\sum_{k'=1}^K f(x, \mathcal{B}^{k'}_i, \tau^{k'}_i)}$. Exchanging the two terms in Eq. (\ref{eq:q_grad}) with Eq. (\ref{eq:first_exp_term}) and (\ref{eq:second_exp_term}), respectively, we get
\begin{align}
    \nabla_{\phi_i}\frac{1}{S}\mathbb{E}_{q_{\psi_i,\phi_i}(\mathcal{B}, \tau)}\left[ \hat{L}^K_i \right] &= \frac{1}{S}\mathbb{E}_{q_{\psi_i,\phi_i}(\mathcal{B}, \tau)}\Big[ \hat{L}^K_i\sum_{k=1}^K\nabla_{\phi_i} \log q_{\phi_i}(\tau_i^{k})\Big]- \\
    & \quad\frac{1}{S}\mathbb{E}_{q_{\psi_i,\phi_i}(\mathcal{B}, \tau)}\Big[\sum_{k=1}^K\tilde{w}^k_i\nabla_{\phi_i}\log \frac{1}{S}\sum_{j=1}^S q_{\psi_j}(\mathcal{B}^k_i| \tau^k_i)q_{\phi_j}(\tau^k_i) \Big].
\end{align}

For $i\neq s$ we may move the gradient operator into the expectation directly and reuse the derivation of Eq. (\ref{eq:second_exp_term}),
\begin{align}
    \nabla_{\phi_i} \frac{1}{S}\sum_{s\neq i}\mathbb{E}_{q_{\psi_s,\phi_s}(\mathcal{B}, \tau)}\left[ \hat{L}^K_s \right] &= \frac{1}{S}\sum_{s\neq i}\mathbb{E}_{q_{\psi_s,\phi_s}(\mathcal{B}, \tau)}\left[ \nabla_{\phi_i}\hat{L}^K_s \right]\\
    &= -
    \frac{1}{S}\sum_{s\neq i}\mathbb{E}_{q_{\psi_s,\phi_s}(\mathcal{B}, \tau)}\Big[\sum_{k=1}^K\tilde{w}^k_s\nabla_{\phi_i}\log \frac{1}{S}\sum_{j=1}^S q_{\psi_j}(\mathcal{B}^k_s| \tau^k_s)q_{\phi_j}(\tau^k_s)\Big]\nonumber.
\end{align}

Considering both cases, we return to Eq. (\ref{eq:grad_i_miselbo_appendix}),
\begin{equation}
\label{eq:self-normalized_gradient_appendix}
\begin{split}
     \nabla_{\phi_i} \mathcal{L}(X;K, S) &= \frac{1}{S}\mathbb{E}_{q_{\psi_i,\phi_i}(\mathcal{B}, \tau)}\Big[ \hat{L}^K_i\sum_{k=1}^K\nabla_{\phi_i} \log q_{\phi_i}(\tau_i^{k})\Big] -\\&\quad
    \frac{1}{S}\sum_{s= 1}^S\mathbb{E}_{q_{\psi_s,\phi_s}(\mathcal{B}, \tau)}\Big[\sum_{k=1}^K\tilde{w}^k_s\nabla_{\phi_i}\log \frac{1}{S}\sum_{j=1}^S q_{\psi_j}(\mathcal{B}^k_s| \tau^k_s)q_{\phi_j}(\tau^k_s)\Big],
\end{split}
\end{equation}
which is the expression for the gradient we need in order to apply the VIMCO estimator (see Sec. \ref{sec:vimco}).

\section{Additional Experimental Results and Implementation Details}

\subsection{The Two-Level Hierarchical Model}
\label{app:twolevelmodel}
\begin{figure}[!t]
    \centering
    \begin{subfigure}{0.4\textwidth}
    \centering
    \includegraphics[width=\textwidth]{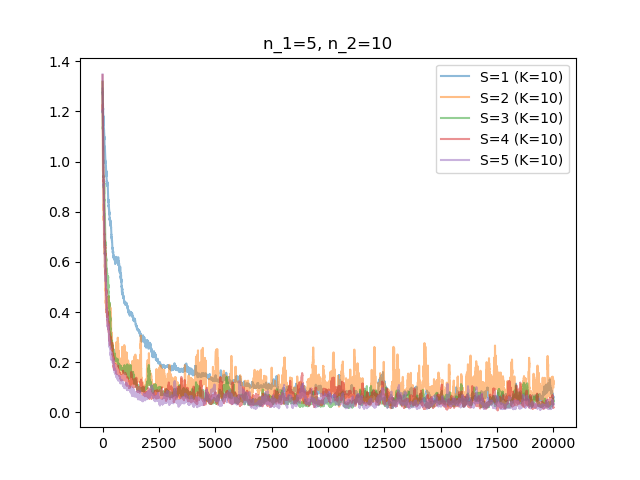}
    \caption{}
    \label{fig:kl_curve-a}
    \end{subfigure}
    ~
    \begin{subfigure}{0.4\textwidth}
    \centering
    \includegraphics[width=\textwidth]{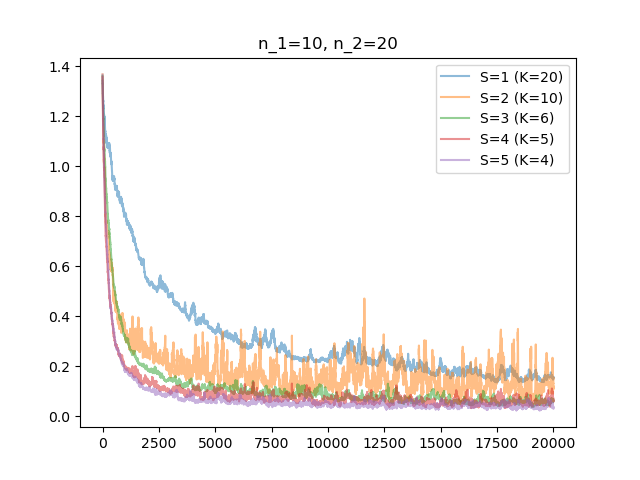}
    \caption{}
    \end{subfigure}
    ~
    \begin{subfigure}{0.4\textwidth}
    \centering
    \includegraphics[width=\textwidth]{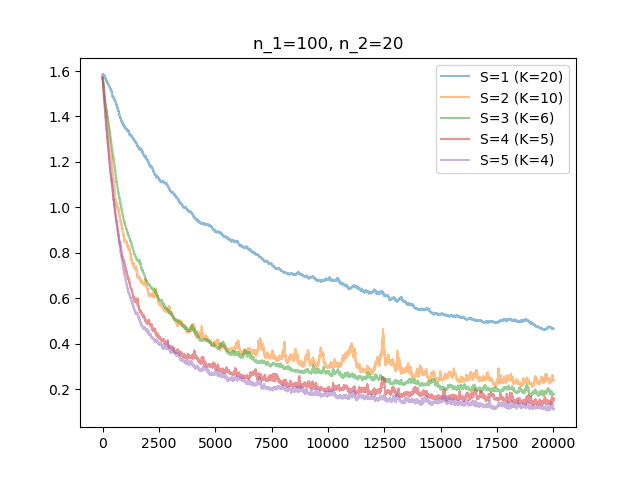}
    \caption{}
    \end{subfigure}
    ~
    \begin{subfigure}{0.4\textwidth}
    \centering
    \includegraphics[width=\textwidth]{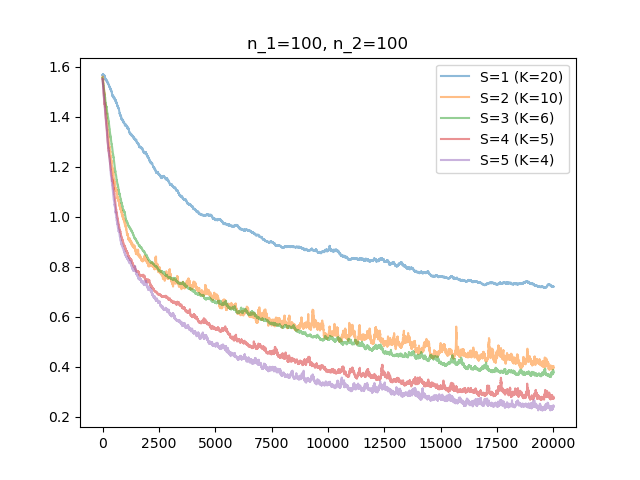}
    \caption{}
    \end{subfigure}
    \caption{KL curves for the two-level hierarchical model using different configurations of $K$, $n_1$ and $n_2$.}
    \label{fig:kl_curves_2}
\end{figure}

Running a grid search for all five algorithms ($S=1,...,5$), using $n_1 = 5$ and $n_2=10$, we found that the learning rates 0.01, 0.1, 0.1, 0.2 and 0.25, respectively, were optimal. That is, these achieved the smallest KL divergences. For larger learning rates, $S=1$ did not converge, or converged to worse KL divergences. The optimal learning rates found in the grid searches were used in all subsequent experiments.

In Fig. \ref{fig:kl_curves_2}, we visualize the KL curves as functions of the number of training iterations. For all configurations of $K$, $n_1$ and $n_2$, the $S=5$ model performs best. The pattern---$S=1$ converges slower and to worse KL divergences than $S>2$---holds also when all models use the same number of importance samples, $K$ (shown in Fig \ref{fig:kl_curve-a}).

\subsection{Visualization Details and More Plots}
\label{app:visualization}

We employed the software as in \cite{whidden2015quantifying}, i.e. rSPR was used to determine the distances between the topologies. Additionally, we adopted the same methodology for cluster creation. This involved assigning the most probable peak to a cluster and subsequently assigning all unassigned trees to the same cluster if their distance from the peak tree was within one standard deviation below the mean distance of all unassigned trees. This iterative process continued until all topologies were assigned, or until eight clusters were reached.

\begin{figure}[!ht]
    \centering
     \begin{subfigure}{0.33\textwidth}\includegraphics[width=\textwidth]{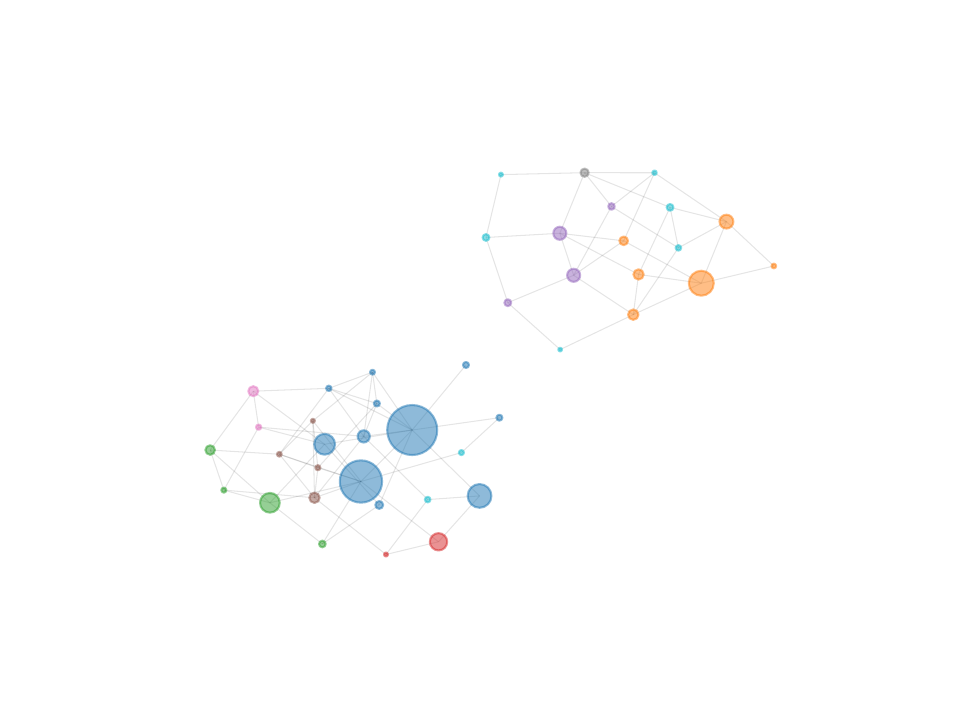}
    \caption{}
    \label{fig:ds1}
    \end{subfigure}~
 \begin{subfigure}{0.33\textwidth}\includegraphics[width=\textwidth]{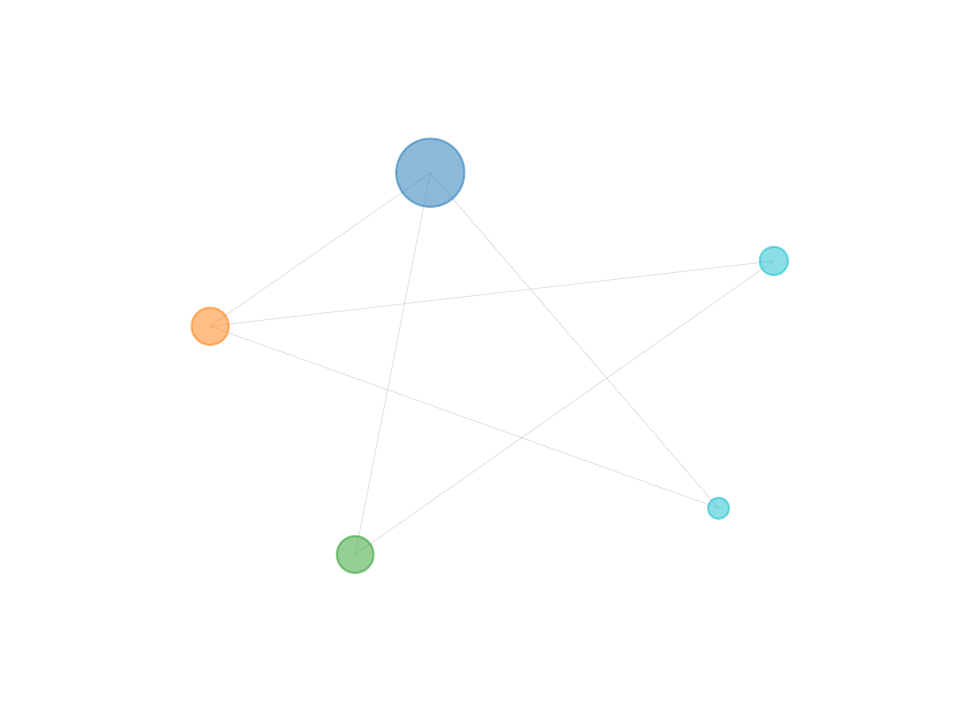}
    \caption{}
    \label{fig:ds2}
    \end{subfigure}~
\begin{subfigure}{0.33\textwidth}\includegraphics[width=\textwidth]{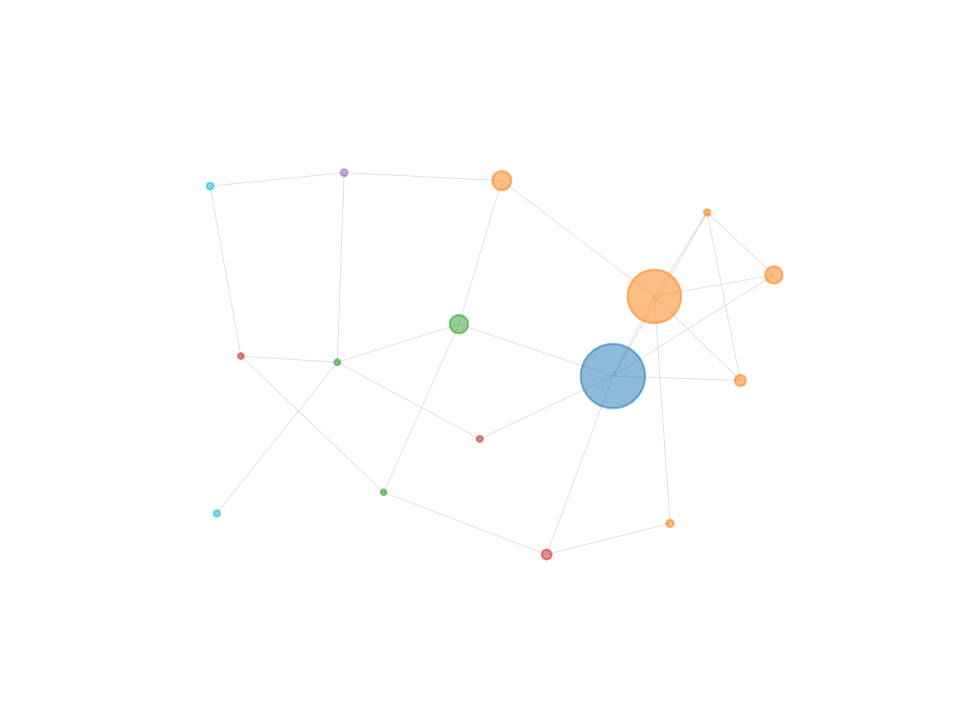}
    \caption{}
    \label{fig:ds3}
    \end{subfigure}
\begin{subfigure}{0.33\textwidth}\includegraphics[width=\textwidth]{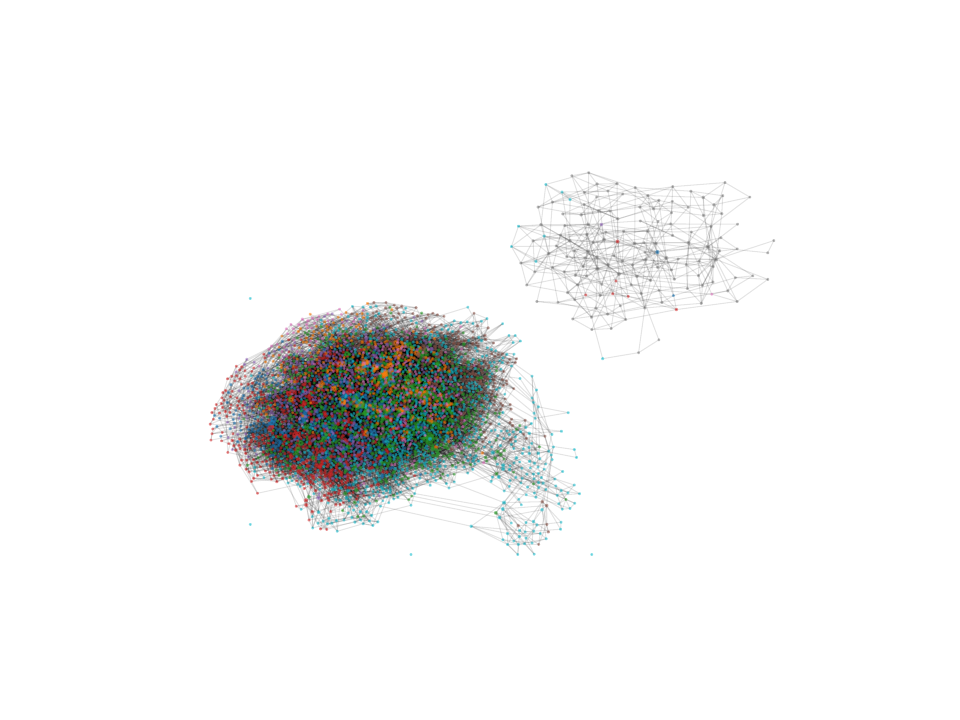}
    \caption{}
    \label{fig:ds5}
    \end{subfigure}
\begin{subfigure}{0.33\textwidth}\includegraphics[width=\textwidth]{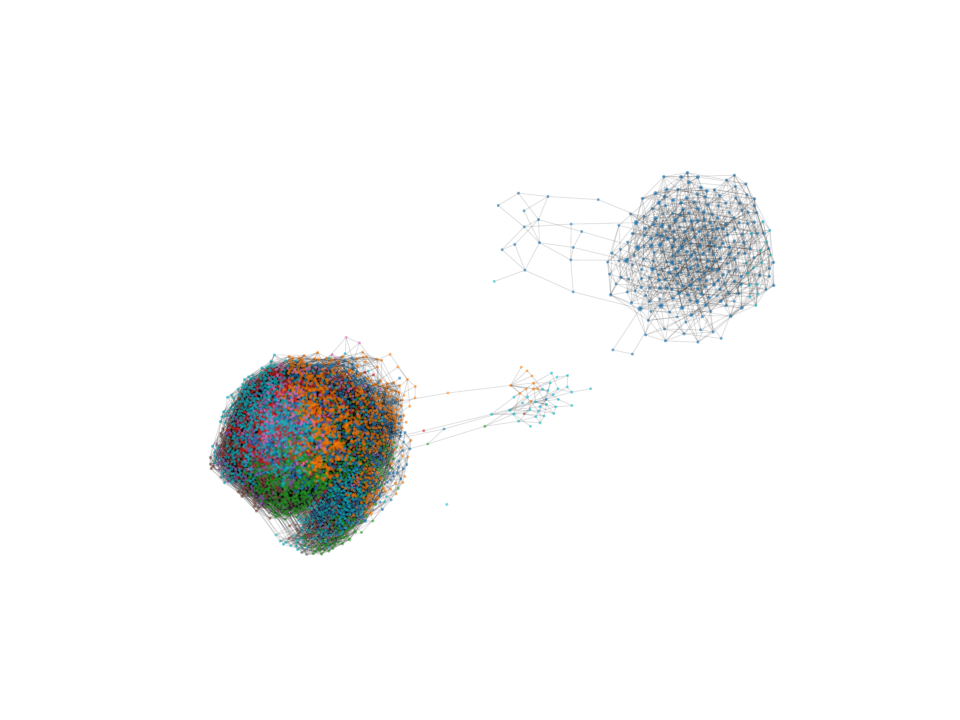}
    \caption{}
    \label{fig:ds6}
    \end{subfigure}
    \caption{Visualization of samples from the tree-topology posterior using a 1,000,000,000 iterations long MCMC run on (a) DS1, (b) DS1, (c) DS3, (d) DS5 and (e) DS6. Nodes represent unique tree-topologies and are colored based on cluster assignments, illustrating the multimodality of the tree-topology posterior. More details in Sec. \ref{sec:background}.}
    \label{fig:mcmc_topology_space2}
\end{figure}

For graph creation, we employ the Graphviz layout known as Scalable Force-Directed Placement (SFDP), in conjunction with the NetworkX library \cite{hagberg2008exploring}. The clusters are represented by different colors, and the size of each node is determined by the normalized sampling frequency in Fig \ref{fig:mcmc_topology_space}, \ref{fig:mcmc_topology_space2}. Moreover, edges between the topologies are only displayed if their distance is exactly one. It's worth noting that the rSPR distance measure is utilized, which counts the number of changes similar to the approach used in the MCMC method.

To ensure that the visualization focused on the most credible information, we imposed a constraint by limiting the nodes to the 95\% most credible set. This ensured that only the most reliable nodes were included. Additionally, to manage computational resources effectively, we set a maximum limit of 4096 nodes for the graphs.

\begin{figure}[!ht]
    \centering
    \begin{subfigure}{0.19\textwidth}
    \centering
    \includegraphics[width=\textwidth]{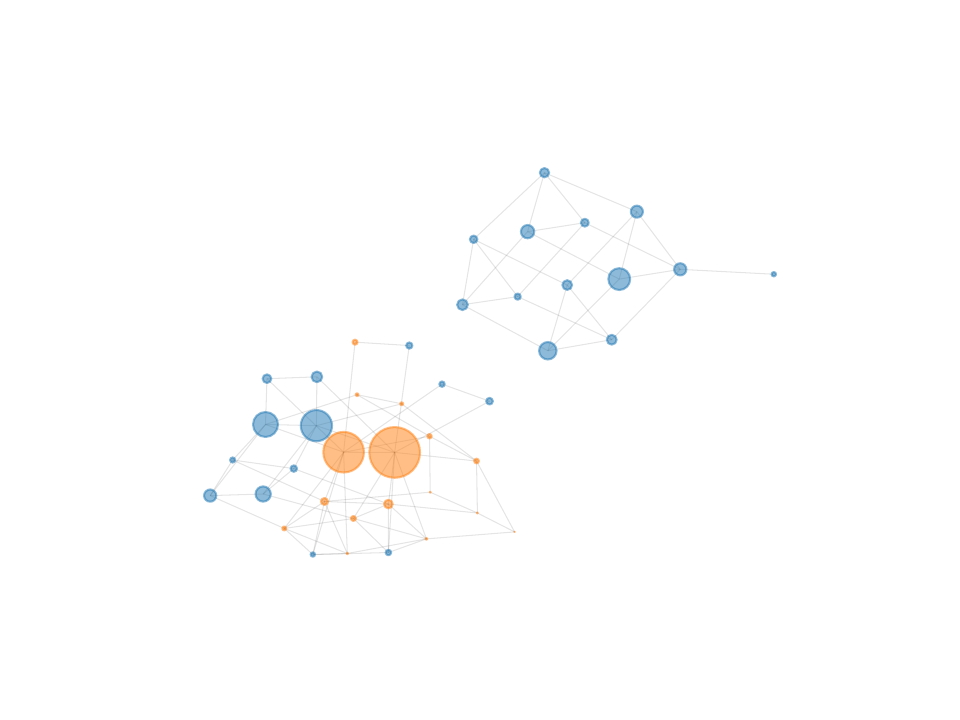}
    \end{subfigure}
    ~
    \begin{subfigure}{0.19\textwidth}
    \centering
    \includegraphics[width=\textwidth]{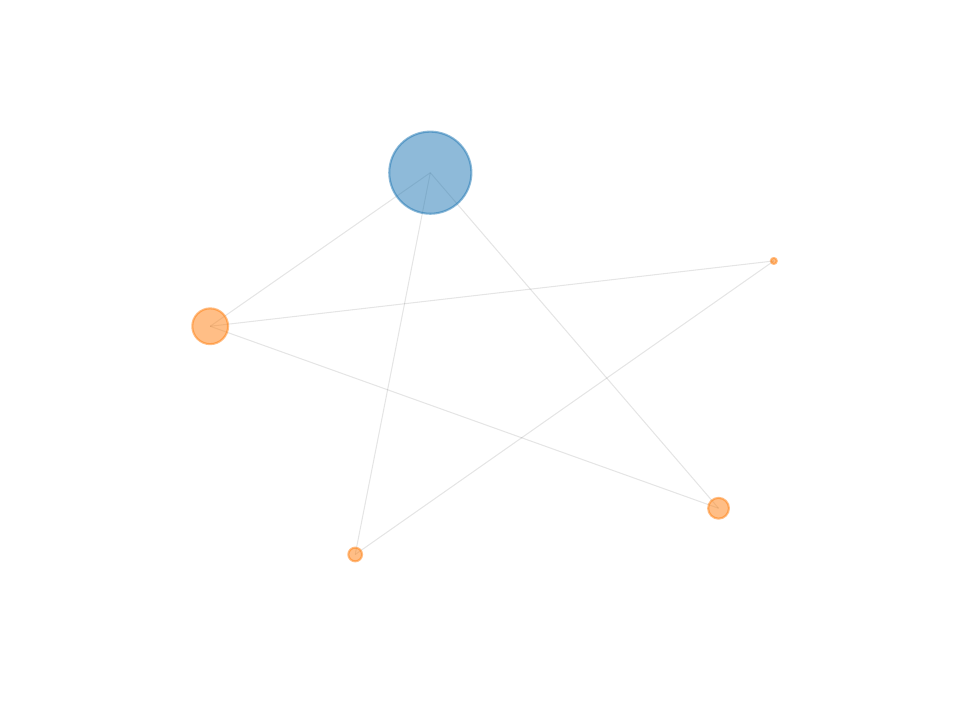}
    \end{subfigure}
    ~
    \begin{subfigure}{0.19\textwidth}
    \centering
    \includegraphics[width=\textwidth]{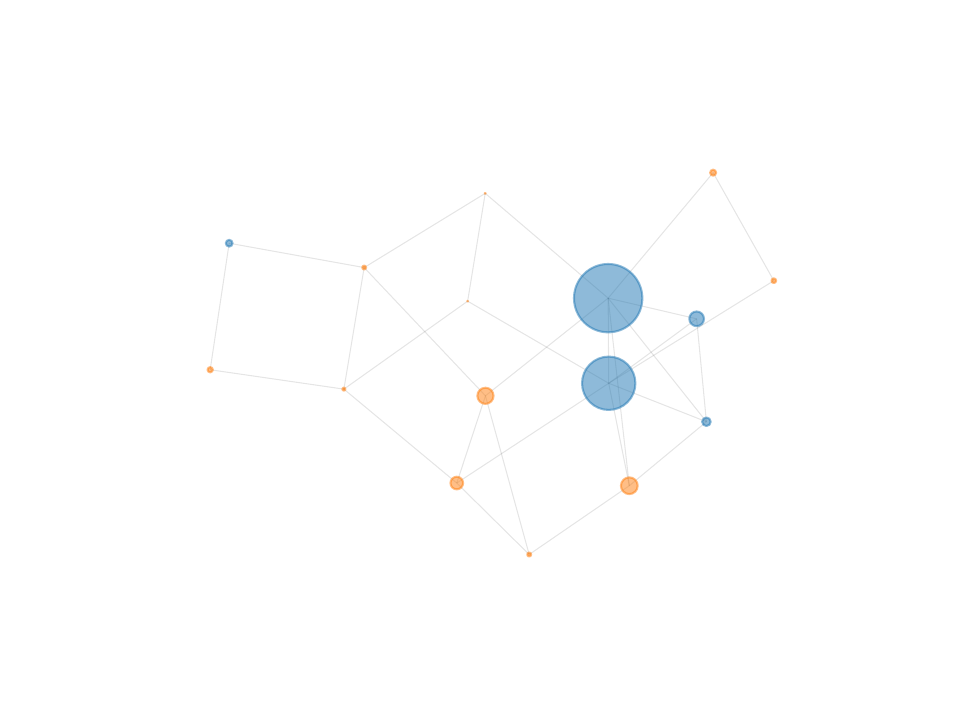}
    \end{subfigure}
    \begin{subfigure}{0.19\textwidth}
    \centering
    \includegraphics[width=\textwidth]{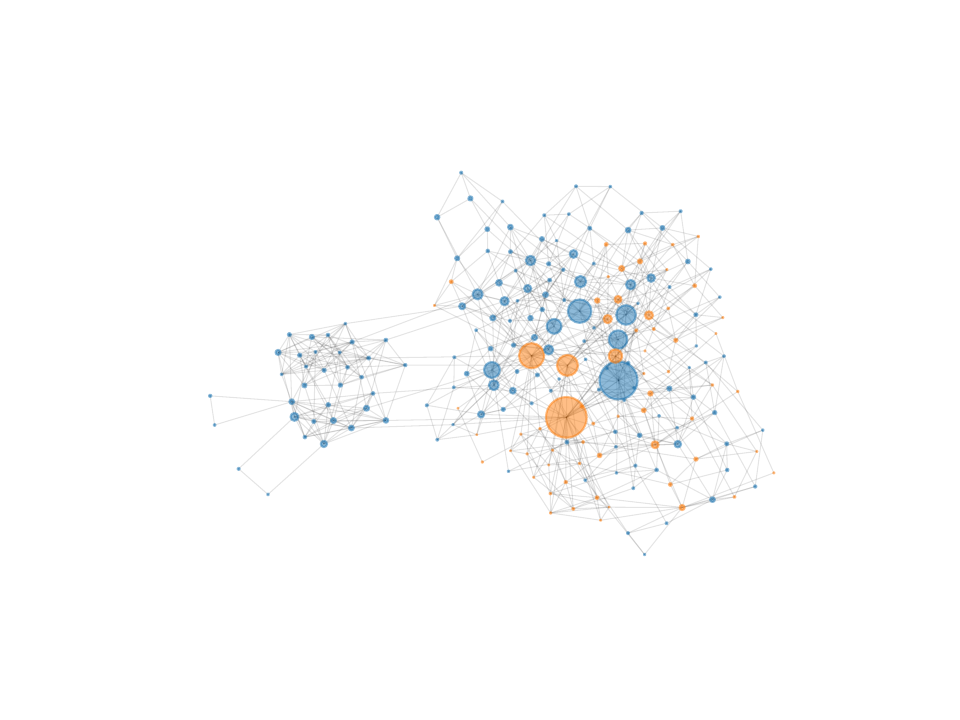}
    \end{subfigure}
    \begin{subfigure}{0.19\textwidth}
    \centering
    \includegraphics[width=\textwidth]{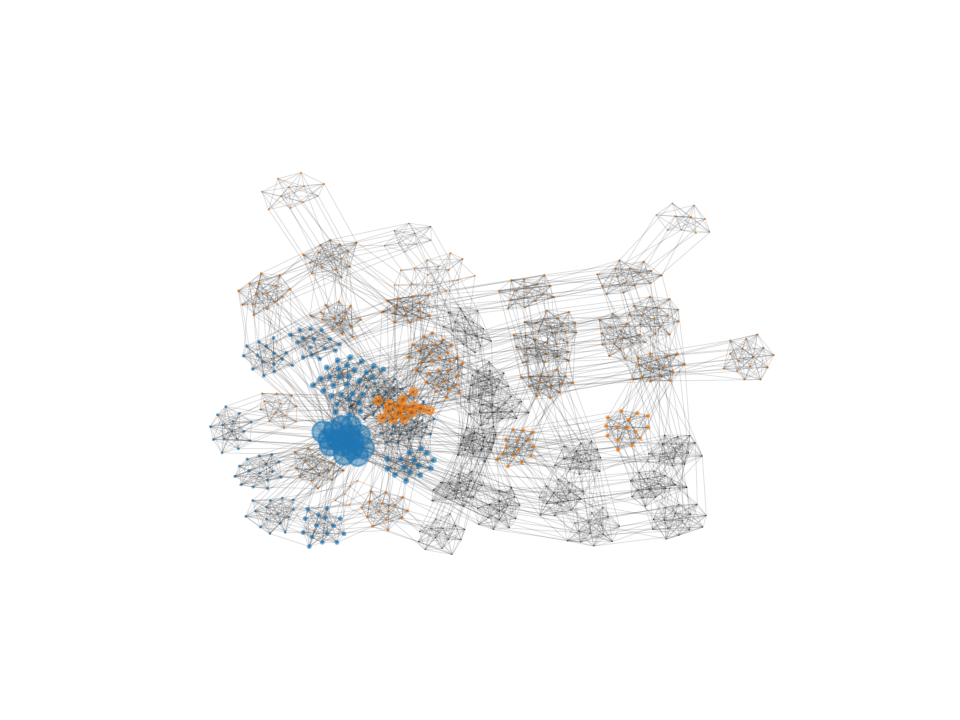}
    \end{subfigure}
    \begin{subfigure}{0.19\textwidth}
    \centering
    \includegraphics[width=\textwidth]{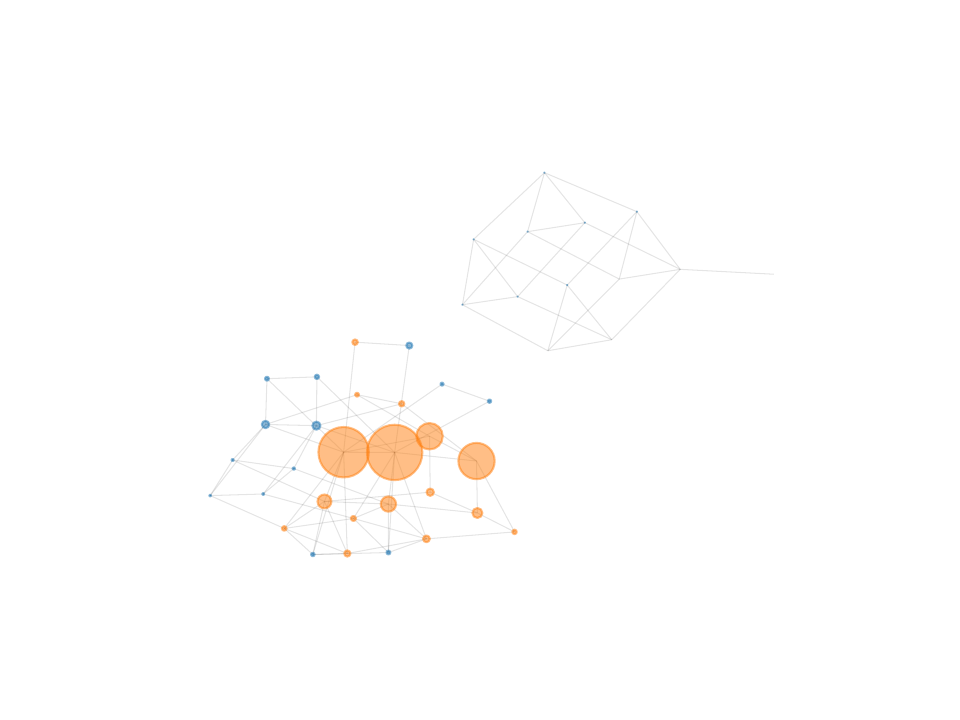}
    \caption{DS1}
    \end{subfigure}
    ~
    \begin{subfigure}{0.19\textwidth}
    \centering
    \includegraphics[width=\textwidth]{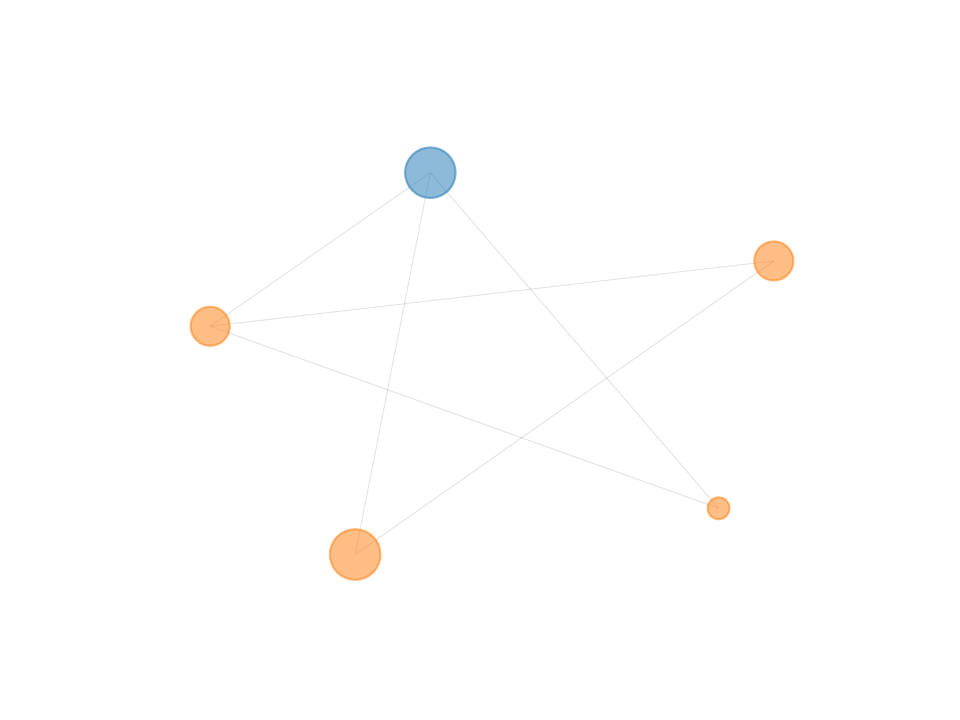}
    \caption{DS2}
    \end{subfigure}
    ~
    \begin{subfigure}{0.19\textwidth}
    \centering
    \includegraphics[width=\textwidth]{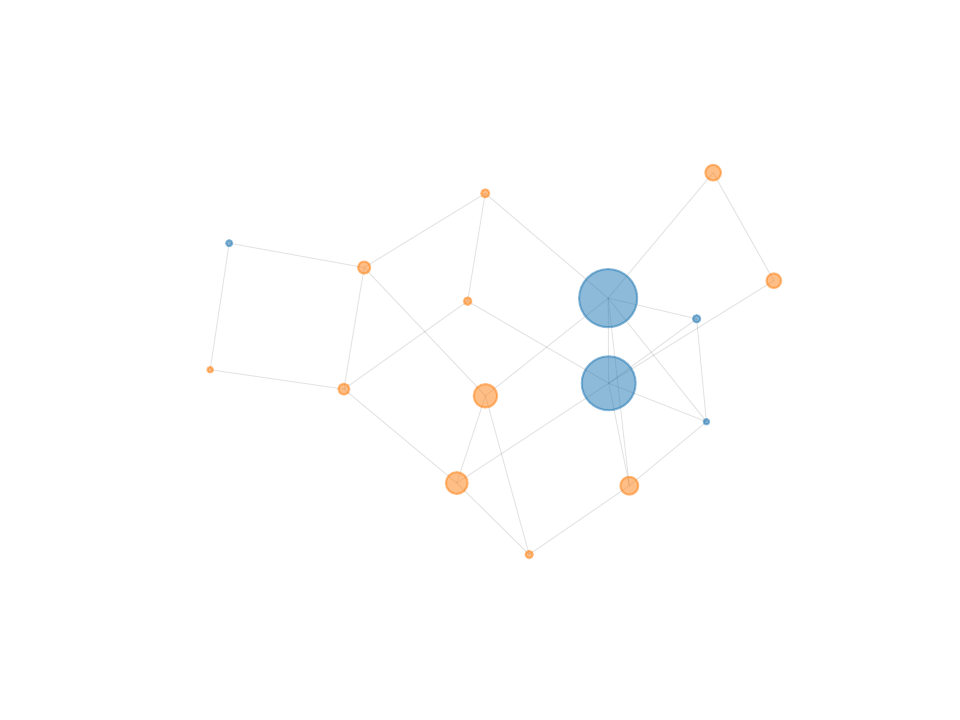}
    \caption{DS3}
    \end{subfigure}
        \begin{subfigure}{0.19\textwidth}
    \centering
    \includegraphics[width=\textwidth]{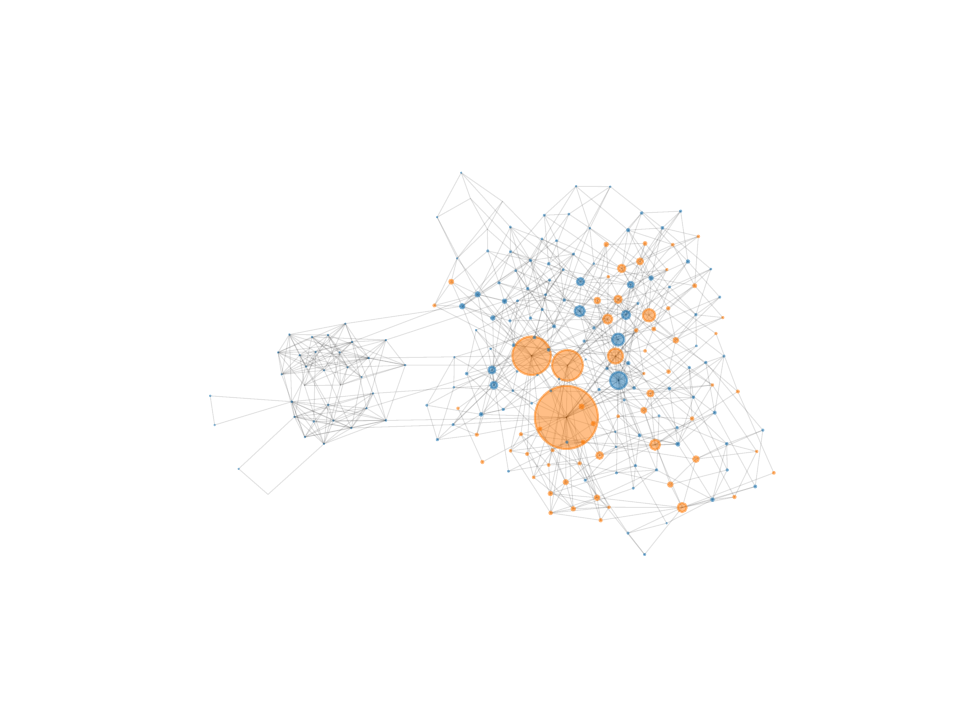}
    \caption{DS4}
    \end{subfigure}
        \begin{subfigure}{0.19\textwidth}
    \centering
    \includegraphics[width=\textwidth]{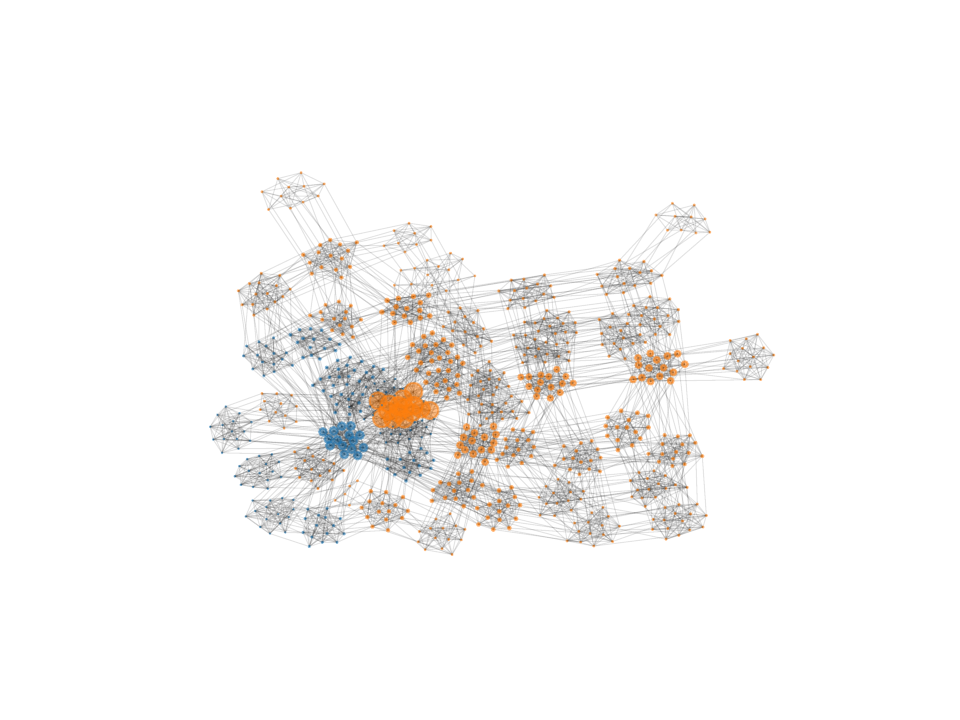}
    \caption{DS7}
    \end{subfigure}
    \caption{Visualization of a uniformly weighted $S=2$-component mixture of SBNs on (a) DS1, (b) DS2 and (c) DS3, (d) DS4 and (e) DS7, where each node corresponds to a unique tree-topology. The upper row shows the distribution of five million sampled tree topologies from the first component, where a node, $\tau$, is colored blue if $q_{\phi_1}(\tau)>q_{\phi_2}(\tau)$, or orange otherwise. Vice versa for the lower row. The size of a node is determined by its sampling frequency, which is why nodes with low frequency appear black. The components clearly spread out, exploring different parts of the space.}
\label{fig:topology_visualization2}
\end{figure}

A similar approach was used for Fig. \ref{fig:topology_visualization}, \ref{fig:topology_visualization2}, with the main difference being that we sampled from the components and displayed the joint set of topologies. The colors in this figure were based on which component sampled the topology the most. This representation was considered an approximation of the posterior.

\section{Limitations}
We use $S$ SBNs and branch length models to form our mixture approximations. This introduces a larger number of model parameters. In our current implementation, the training time was prolonged as we did not parallelize the parameter updates of the parameters of the mixture components. This can, however, be done, in order to heavily decrease training time.

Additionally, using shared parameters for the mixture components can also be utilized, if the practitioner is running on a limited memory budget. However, with modern compute engines and laptop computers, this is seldomly an issue. Nonetheless, devising clever modifications to reduce the number of parameters and training times for mixtures in black-box VI is an exciting future research field, out of the scope of this work.

\section{Broader Impact}
Bayesian phylogenetic inference algorithms are crucial for researchers to reason about uncertainty in their evolutionary findings. Variational inference algorithms provide a compelling alternative to MCMC-based algorithms as a parametric approximation is obtained. This implies that VI, and VBPI specifically, can be used in settings where the applcation of MCMC is less straightforward, for instance in out-of-distribution detection, or evaluation on held-out data. Also, as we have shown in our experiments in this paper, model evaluation can be more robust when using VI over MCMC, resulting in smaller variance of the estimator of the marginal log-likelihood. This is an important feature for downstream tasks. 

\section{Compute Infrastructure}
Most computations have been conducted on an AMD EPYC 7742 where two cores have been used per run. Final runtimes are shown in Table \ref{tab:compute}. The table shows the joint run time for both training and testing. Also worth noting is that multiple mixture components also multiply the number of particles, so the majority of the time increase is due to the Felsenstein pruning algorithm for the likelihood model evaluation, which grows linearly. 

Finally, and crucially, the code used was not optimized for run time, and so a wall-clock time is not an apt metric for comparisons.

\begin{table}[ht!]
    \centering
    \caption{Compute time reported in minutes for fitting the model as well as evaluating the marginal likelihood and continuous estimate of ELBO while training  every 5000 iterations using 1000 samples with a single particle}
    {\tiny
    \tabcolsep1.1pt
    \begin{tabularx}{\linewidth}{@{}X|XXXXXXXX@{}}
\hline
Data & DS1 & DS2 & DS3 & DS4 & DS5 & DS6 & DS7 & DS8 \\\hline
VBPI                                          & 735.4   & 813.95  & 962.67  & 1087.98 & 1273.57 & 1299.55 & 1606.03 & 1660.18 \\
Mix\textsubscript{$S=2$}     & 1984.47 & 2172.18 & 2598.17 & 2888.58 & 3322.77 & 3389.5  & 4127.52 & 4234.98 \\
Mix\textsubscript{$S=3$}     & 3224.57 & 3425.35 & 4056.07 & 4493.87 & 5614.58 & 5604.43 & 6705.18 & 7004.28 \\
VBPI-NF                                       & 900.32  & 953.8   & 1150.78 & 1278.27 & 1585.87 & 1523.02 & 1869.23 & 1974.17 \\
Mix\textsubscript{NF, $S=2$} & 2385.38 & 2442.75 & 2893.15 & 3300.35 & 3800.88 & 3728.75 & 4647.48 & 4754.22 \\
Mix\textsubscript{NF, $S=3$} & 3669.8  & 3939.73 & 4576.08 & 5180.75 & 6206.52 & 6220.4  & 7568.85 & 7776.37
\end{tabularx}
    }
    \label{tab:compute}
\end{table}

\end{document}